\documentclass{article}

\usepackage{PRIMEarxiv}
\usepackage[utf8]{inputenc} % allow utf-8 input
\usepackage[T1]{fontenc}    % use 8-bit T1 fonts
\usepackage{hyperref}       % hyperlinks
\usepackage{url}            % simple URL typesetting
\usepackage{booktabs}       % professional-quality tables
\usepackage{amsfonts}       % blackboard math symbols
\usepackage{nicefrac}       % compact symbols for 1/2, etc.
\usepackage{microtype}      % microtypography
\usepackage{lipsum}
\usepackage{fancyhdr}       % header
\usepackage{graphicx}       % graphics
\usepackage{amsmath}
\usepackage{subcaption}
\usepackage{wrapfig}
\usepackage{natbib}
\usepackage{comment}
\usepackage{rotating}
\usepackage[table]{xcolor}
\usepackage{soul}
\sethlcolor{green}
\graphicspath{{resources/}}     % organize your images and other figures under media/ folder

\title{\Large A Practical Causal Framework for Autonomous Mobile Robots:\\From Simulation to Real-World Human Environments}

\author{
 Luca Castri \\
  University of Padua, Italy\\
  \texttt{lucacastri94@gmail.com} \\
   \And
 Gloria Beraldo \\
  University of Padua, Italy\\
  National Research Council of Italy\\
  \texttt{gloria.beraldo@cnr.it} \\
  \And
 Nicola Bellotto \\
  University of Padua, Italy\\
  \texttt{nbellotto@dei.unipd.it} \\
}

\begin{document}
\maketitle

\begin{abstract}
Causal inference has emerged as a powerful paradigm enabling robots to make safe, efficient decisions by reasoning about the ``why'' behind events. However, its real-world application is hindered by the difficulty of collecting causal data, especially in human-populated environments where training takes time, while users expect immediate operation. To this end, we tackle the challenge of transferring simulation-learnt causal relations to the real world with minimal effort. While retaining the invariant high-level causal structure, we empirically demonstrate the system adapts efficiently to real-world situations by updating only the causal model parameters with a limited set of new observations. Our results show that the parameter update enables rapid sim-to-real transfer. The framework runs efficiently onboard of a mobile robot and significantly outperforms a non-causal baselines navigating successfully and safely in complex human-shared scenarios.
\end{abstract}

% keywords can be removed

\section{INTRODUCTION}
Deploying autonomous mobile robots in human-populated environments (e.g., warehouses, hospitals, airports, malls) requires safe and efficient interaction with humans. Understanding how robot actions causally affect their surroundings enables better action selection, motivating methods that reason about causal relationships in the environment and human-robot interactions.
%
% Literature review
Accordingly, the robotics community has increasingly embraced causal approaches in various fields~\cite{hellstrom2021relevance}, including robot manipulation~\cite{Lee2022,cannizzaro2025cobra, castri2024candoit}, explainable robotics and HRI~\cite{diehl2022did,love2024towards}, spatial interactions~\cite{castri2024experimental}, planning~\cite{cannizzaro2023car}, and robot understanding~\cite{edstrom2023robot}. 
% Gap
However, transitioning these algorithmic frameworks from theoretical formulation to real-world deployment remains a challenge. Even in purely robotic scenarios, this process is hindered by the need for extensive and time-consuming data collection~\cite{zhao2020sim}. This difficulty is amplified in human-populated settings, where long-term data collection is often impractical.

Despite these constraints, users typically expect robots to be immediately functional. To meet this operational demand, a robust strategy must leverage simulation for extensive data collection and subsequent causal discovery, while enabling rapid adaptation to real-world dynamics. This necessitates a modular, end-to-end causal inference framework---spanning from raw sensor data acquisition and structure learning directly to online reasoning. Crucially, such a framework must be capable of efficient sim-to-real transfer, retaining the high-level logic learnt in simulation while seamlessly updating its parameters to the physical environment.
\begin{figure}[t]
    \centering
  \includegraphics[trim={0cm 0cm 0cm 0cm},clip,width=0.7\columnwidth]{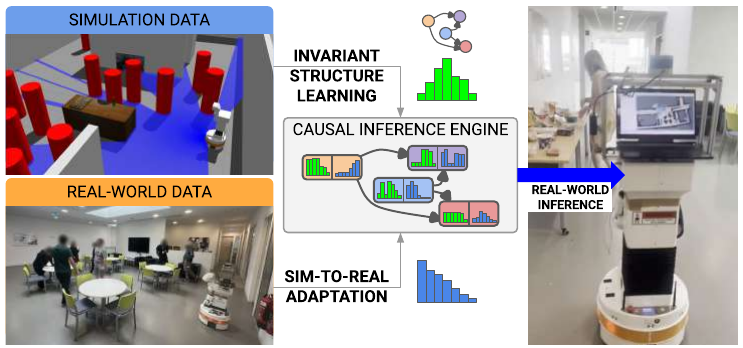}
  \caption{Framework overview.}
  \label{fig:intro}
\end{figure}
Prior work~\cite{castri2026causality} laid the groundwork for our approach with a modular, causality-enhanced decision-making framework supporting both causal discovery and reasoning. However, the authors evaluated task execution efficiency and human safety only in simulation, leaving real-world deployment and validation unaddressed. This paper bridges the gap between simulated and real applications by proposing an improved framework with a data-efficient adaptation process that tunes the initial simulation-learnt causal model to work equally well on a physical mobile robot in human environments (Fig.~\ref{fig:intro}). 
We validate this approach in two real-world scenarios -- a poster session and a social event -- proving that the causal framework operates as expected onboard of a robot for solving challenging navigation tasks, outperforming a non-causal alternative while maintaining low inference time. 
Our main contributions include:
\begin{itemize} 
    \item A new data-efficient sim-to-real transfer that exploits structural causal invariance to adapt the framework via rapid parameter update on limited real-world data.
    \item The first real-world deployment and validation of an end-to-end causality-based framework applied to robot navigation tasks in human environments, made publicly available\footnote{{{\fontsize{6.5pt}{7.5pt}\selectfont
\url{https://lcastri.github.io/causal-sim2real/}}}}\label{foot:website} for reproducibility and future applications.
    \item An experimental evaluation across challenging simulated and real-world group encounters, focusing on task success, operational~safety and inference time.
\end{itemize}
%
%A webpage containing source code and videos of the real-world experiments is available online\footnote{{{\fontsize{6.5pt}{7.5pt}\selectfont
%\url{https://anonymous.4open.science/r/causal-sim2real-7E0D/}}}} \label{foot:website}.

% Paper structure
%The rest of this paper is structured as follows: Section~\ref{sec:related} reviews applications of causality in robotics; Section~\ref{sec:approach} details the causality-based decision-making framework and its real-world adaptability; Sections~\ref{sec:exp-setup}, \ref{sec:exp-sim}, and \ref{sec:exp-real} present the experimental setup and evaluate the proposed approach in simulated and real environments. Finally, Section~\ref{sec:conclusion} summarises our achievements and outlines future work.

\section{RELATED WORK}\label{sec:related}
In recent years, the synergy between robotics and causality has gained increasing attention~\cite{hellstrom2021relevance}. However, learning causal models (i.e. {\em causal discovery}) requires extensive datasets that are %prohibitive to collect on physical hardware, due to safety and time constraints, in human-populated environments. 
%Researchers can rely on simulation for data generation, but transferring these models to the real world is often hindered by mismatches in environment and human dynamics.
costly and potentially unsafe to collect on physical robots in human-shared environments. Simulation offers a practical alternative, but transferring learned models to the real world is hindered by mismatches in environment and human dynamics.
%Standard approaches typically fall into two categories: \textit{domain randomisation}, which varies simulation parameters (e.g., friction, lighting) to make the robot's behaviour robust to real-world variations~\cite{tobin2017domain}, and \textit{domain adaptation}, which attempts to map feature representations from the source (simulation) to the target (real world) domain~\cite{tzeng2017adversarial}. 
Standard transfer approaches include \textit{domain randomisation}, which varies simulation parameters to improve robustness to real-world variations~\cite{tobin2017domain}, and \textit{domain adaptation}, which aligns feature representations between simulation and reality~\cite{tzeng2017adversarial}.
%While effective for perception and reactive control, these methods often result in ``black-box'' policies that latch onto spurious correlations---such as linking specific environmental patterns directly to final task success---rather than identifying the true underlying causal mechanisms governing the environment. Additionally, when deployed in novel physical spaces, these policies lack interpretability and require significant real-world data to be fine-tuned.
While effective for perception and reactive control, these methods often learn ``black-box'' policies based on spurious correlations rather than underlying causal mechanisms. Their lack of interpretability and reliance on real-world data limit adaptation to novel environments.

%To address these limitations, recent research has explored causality as a robust mechanism for transfer learning. The central hypothesis is that, while statistical correlations (e.g., background textures or lighting) may change between domains, the underlying \textit{causal mechanisms} (e.g., physics, collision dynamics) remain invariant~\cite{scholkopf2021toward,rahimi2025sim}.
Causality has therefore emerged as a promising mechanism for robust transfer learning. The key assumption is that, although statistical correlations may change across domains, the underlying \textit{causal mechanisms} remain invariant~\cite{scholkopf2021toward,rahimi2025sim}. In our setting, this means that causal relationships between semantically meaningful variables, reflecting underlying physical and interaction mechanisms, are preserved across simulation and reality, while their conditional distributions may change. This assumption is not specific to our approach, but is also adopted by established causal learning methods such as Invariant Risk Minimization (IRM)\cite{arjovsky2019invariant} and causal reinforcement learning\cite{huang2021adarl}, which seek to identify relationships that remain stable across environments, with applications to motion forecasting~\cite{liu2022towards}, intention prediction~\cite{hu2022causal}, and simulation-to-real adaptation~\cite{rahimi2025sim}.
%Leveraging this principle, several works have applied ``causal invariance'' to reinforcement learning, for instance to learn invariant features that hold true across environments~\cite{huang2021adarl,arjovsky2019invariant}, while others have employed causal representations to identify specific model components requiring adaptation~\cite{liu2022towards, hu2022causal,rahimi2025sim}. 
%Leveraging this principle, representation learning methods like Invariant Risk Minimization (IRM)~\cite{arjovsky2019invariant} and causal reinforcement learning frameworks such as AdaRL~\cite{huang2021adarl} force models to identify features and transition dynamics that remain reliable across multiple environments, thereby mathematically discarding spurious domain-specific correlations. 
%Within the context of autonomous navigation, causal representation learning has been successfully employed to improve pedestrian motion forecasting~\cite{ liu2022towards} and vehicle intention prediction~\cite{hu2022causal} by disentangling invariant dynamic behaviours from environment-specific styles. Recent metric learning approaches have demonstrated the potential of aligning simulation and real-world data distributions causally to improve human-aware interaction models~\cite{rahimi2025sim}.
Previous work showed, in a more limited scenario, that causal models learned independently from scratch in simulation and in the real world exhibited consistent structures~\cite{castri2024ros}. Building on these findings, the proposed approach explicitly transfers the graphical—and therefore explainable—causal model from simulation to the real world. By retaining the graph topology and calibrating only the required parameters, it enables rapid, data-efficient adaptation from limited real-world observations, avoiding extensive retraining.
%Unlike previous approaches, which often use implicit causal relations to regularise neural networks, the proposed approach explicitly transfers graphical---and therefore explainable---causal models from simulation to the real world. By retaining the invariant graph topology and updating only the required parameters, our solution enables rapid, data-efficient adaptation with limited real-world observations, avoiding extensive retraining.%\\[0.1cm]
%

%\noindent\textbf{Causal Robotics:} While causal inference is increasingly applied in robotics, existing research typically treats \textit{discovery} and \textit{reasoning} as isolated tasks.
%In mobile robotics, several approaches for causal discovery of such models from time-series data, such as PCMCI~\cite{runge_detecting_2019} and its extended version that includes a feature selection mechanism, F-PCMCI~\cite{castri2023enhancing}, have been used to forecast human spatial behaviours. Notably, ROS-Causal~\cite{castri2024experimental} integrates these state-of-the-art methods into the ROS ecosystem, thereby enabling onboard discovery directly from robot sensor data. However, previous research was generally limited to causal discovery, lacking online reasoning capabilities.
In robotics, causal discovery methods such as PCMCI~\cite{runge_detecting_2019} and F-PCMCI~\cite{castri2023enhancing} have been used to forecast human spatial behaviour from time-series data. ROS-Causal~\cite{castri2024experimental} further enables onboard causal discovery from robot sensor data. However, these works primarily focus on causal discovery and do not address online causal reasoning. Conversely, causal reasoning has been incorporated into POMDPs for safe planning~\cite{cannizzaro2023car}, probabilistic programming for uncertain manipulation~\cite{cannizzaro2025cobra}, and proactive guidance~\cite{love2024towards}.
%Other works have leveraged causal reasoning to enhance decision-making in complex robotic tasks. For instance, explicit causal models have been integrated with POMDPs to enable safe planning in confounded environments~\cite{cannizzaro2023car}, embedded into probabilistic programming frameworks to manage uncertainty during robot manipulation~\cite{cannizzaro2025cobra}, and utilised to generate proactive navigation strategies and explanations in human-robot interaction scenarios~\cite{love2024towards}.
However, these approaches typically rely on pre-defined, static causal structures, rather than learning them from sensor data.

An end-to-end real-world application that learns causal models and updates their parameters directly from onboard sensors for online decision-making is currently missing. Building on the simulation-based architecture in~\cite{castri2026causality}, this work presents the first real-world deployment of such an integrated framework, validating its feasibility and effectiveness for human-aware robot navigation.

\section{CAUSAL ROBOT FRAMEWORK}\label{sec:approach}
\begin{figure}[t]
\vspace{1.5mm}
    \centering
    \includegraphics[trim={0cm 0cm 0cm 0cm},clip,width=0.6\textwidth]{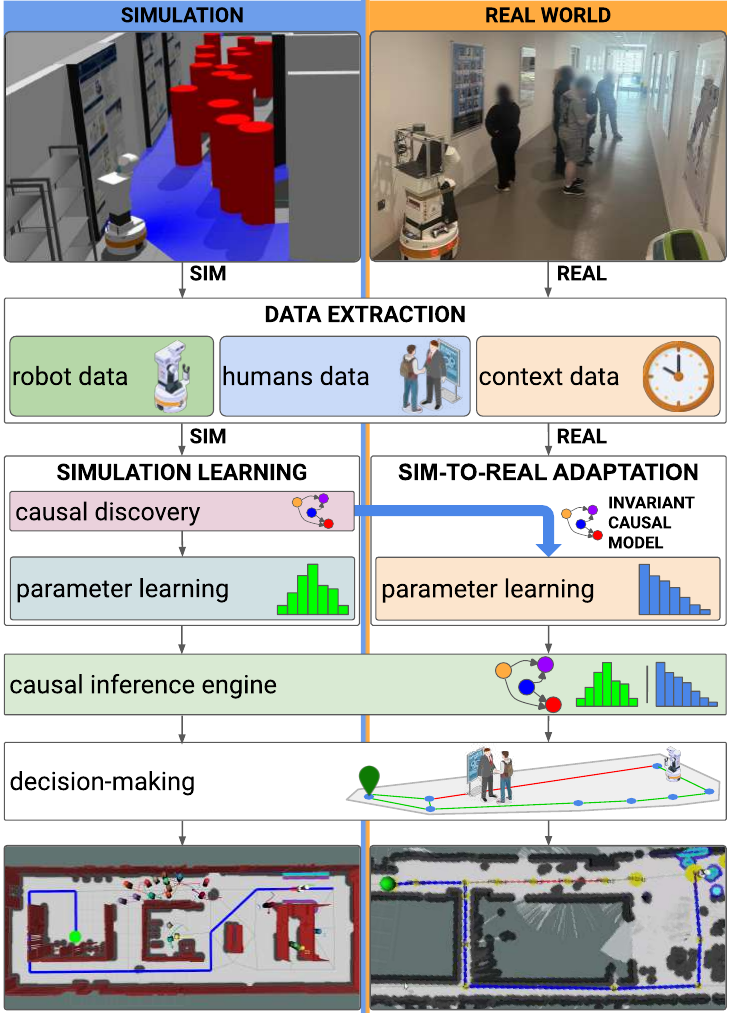}
    \caption{Causal framework: A rapid sim-to-real adaptation calibrates only the conditional probability parameters (green vs. blue histograms), preserving the causal model learned in simulation and enabling efficient robot deployment.}
    \label{fig:approach}
\end{figure}

%The methodology for simulation-based learning and real-world adaptation 
The proposed simulation-to-real approach is shown in Fig.~\ref{fig:approach}. It includes a two-stage process: first, the causal model is learnt and validated by simulated navigation experiments; second, the framework is adapted to the real world via parameter calibration and deployed on a physical robot.
\subsection{Causality-Enhanced Decision-Making}\label{subsec:framework}
The system described in~\cite{castri2026causality} and adopted in this work enables autonomous robots to learn and reason about cause-and-effect relations to take decisions in dynamic, human-shared environments. Its core causal inference engine allows the robot to balance various objectives---such as minimising battery consumption while maximising human safety---by inferring how contextual factors and robot actions causally affect the human behaviour and the robot's internal dynamics. The architecture consists of three pipelines: %\textit{data extraction}, \textit{learning}, and \textit{inference}.

\textbf{Data Extraction:}~This pipeline extracts raw time-series data on robot state (e.g., speed, battery level), human spatial behaviours (e.g., crowd density at specific waypoints over time), and environmental context (e.g., time slot, waypoint). To ensure computational efficiency and compatibility with discrete causal algorithms, the data undergoes two pre-processing steps: Nyquist-Shannon subsampling, and quantile-based discretisation for suitable inference.

\textbf{Learning:}~This pipeline constructs the causal inference engine via \textit{causal discovery}, powered by CausalFlow\footnote{{\fontsize{6.5pt}{7.5pt}\selectfont CausalFlow: \url{https://github.com/lcastri/causalflow}\label{foot:causalflow}~\cite{castri2024candoit}}}, and \textit{parameter learning}, where Maximum Likelihood Estimation (MLE) is used to learn the Conditional Probability Distributions (CPDs) for each node of the identified structure.

\textbf{Inference:}~This final pipeline drives real-time decision-making via causal interventions (e.g., ``\textit{What if I travel to the goal now at speed $s$?}''). The engine applies do-calculus~\cite{pearl2009causality} to estimate the effects of these actions, embedding them into an $\text{A}^*$ heuristic function, as detailed in Sec.~\ref{sec:causal-heuristic}.

\subsection{Simulation-Based Learning}\label{subsec:simulation-learning}
%To facilitate rapid deployment, we leverage causal structural invariance~\cite{scholkopf2021toward,rahimi2025sim}, assuming that high-level causal mechanisms %(e.g., time and location $\rightarrow$ people density $\rightarrow$ velocity $\rightarrow$ battery consumption)
%in simulation remain valid also in the real world, while its parameters (e.g., congestion probabilities) are environment-dependent.
To facilitate rapid deployment, we leverage the notion of causal structural invariance, in line with prior work~\cite{scholkopf2021toward,rahimi2025sim}. Specifically, we consider high-level causal mechanisms %(e.g., time and location $\rightarrow$ people density $\rightarrow$ velocity $\rightarrow$ battery consumption)
to be relatively stable across simulation and real-world, while their parameters (e.g., congestion probabilities) may vary across environments. 

To learn this invariant structure, we used PeopleFlow\footnote{{\fontsize{6.5pt}{7.5pt}\selectfont PeopleFlow: 
\url{https://github.com/lcastri/PeopleFlow}~\cite{castri2026causality}}}\label{foot:peopleflow}, our Gazebo-based social simulator~\cite{castri2026causality}, which models human-robot spatial interactions based on context (e.g. time, location, people density), alongside the robot's internal dynamics, including battery drain adjusted for speed and obstacles.%, as shown in Fig.~\ref{fig:hypothetical-cm}.

%It models the battery level change $L_t$ as a function of velocity $S_t$, charging status $C$, and unexpected static obstacles $O$:
%\begin{equation}\label{eq:delta-rb}
%L_t =
%\begin{cases}
%-\Delta_t \Big( K_{s} + K_{d} \cdot S_t \Big) & C = 0, O = 0\\
%-\Delta_t \Big( K_{s} + K_{d} \cdot S_t \Big) K_{o} & C = 0, O = 1\\
%+\Delta_t \cdot K_{c} & C = 1
%\end{cases}
%\end{equation}
%where $\Delta_t$ is the time step, $K_s$ and $K_d$ are the static and dynamic consumption coefficients, and $K_o$ is a penalty coefficient accounting for the increased energy drain caused by obstacle avoidance maneuvers. By combining realistic social crowds with physical robot constraints, PeopleFlow generates the high-variance data required for causal discovery.\\[0.1cm]
%

%
\subsection{Real-World Adaptation via Data-Efficient Update}\label{subsec:adaptation} 
%
%To transfer the simulation-learnt causal model to the real world, we adopt a fine-tuning strategy to learn its parameters under the causal invariance assumption previously mentioned. 
%
%Learning a causal structure from scratch demands high-variance datasets covering diverse contextual situations (e.g., human behaviours across different time slots and tasks). Collecting such comprehensive datasets in the real world is often impractical and time-consuming, while simulation can typically generate data much more efficiently.
%
To transfer the simulation-learnt causal model to the real world, we fine-tune its parameters under the previously discussed causal invariance assumption. Learning the causal structure from scratch would require extensive real-world data across diverse contexts, which is costly and time-consuming. Thus, since the underlying physical and operational mechanisms remain largely invariant, the adaptation process becomes a parametric problem, updating the CPDs with minimal real-world data. In practice, we equip the physical robot with the (invariant) causal graph learnt in simulation, and update its CPDs via MLE using limited real-world data.  
The robot is deployed for a brief observational phase to record real-world contextual drivers (e.g., location, time) and their resulting effects on system variables (e.g., actual crowd density). As our framework employs discrete tabular CPDs and the causal dependencies are already fixed by the invariant graph, the MLE procedure computes the empirical conditional frequencies from this new data. Specifically, for each node $X_i$ and its parent set $Pa(X_i)$, the updated parameters $\hat{P}(X_i \mid Pa(X_i))$ are directly estimated from the observed real-world data. For instance, while the simulation dictates that a specific waypoint at a particular time slot influences crowd density, this update rapidly calibrates the precise probability of congestion based on the physical venue's actual crowd behaviour. This avoids data- and computationally-expensive structure learning, calibrating only the parameters across all variables to reflect the probability distributions of the physical environment.

\section{EXPERIMENTAL SETUP}\label{sec:exp-setup}
%This section evaluates the proposed causality-enhanced framework through a two-stage experimental process. First, we validate the framework's navigation performance in simulation, and demonstrate the causal model's superiority over a standard Bayesian network for inference tasks. Second, we deploy the invariant structure on a physical TIAGo robot to assess whether the framework efficiently adapts to the real world via a light-data parameter update, and whether it improves navigation in dynamic human environments compared to a non-causal baseline. Videos of the real-world experiments are publicly available online\footref{foot:website}.

This section details the experimental setup adopted in our two-stage evaluation strategy (Secs.~\ref{sec:exp-sim}~and~\ref{sec:exp-real}).\\[0.1cm]
\noindent\textbf{Objective:} To evaluate the framework's navigation performance in simulation and reality, alongside its sim-to-real adaptability and real-time viability on a TIAGo robot.\\[0.1cm]
\noindent{\textbf{Scenarios:} Four social scenarios across two environments:
\begin{itemize}
    \item \textbf{Conference:} poster session (S1), with queues forming along a main corridor; buffet event (S2), where groups gather around food stations.
    \item \textbf{Airport:} passenger crowds queuing at baggage reclaim belts upon flight arrivals (S3 and S4).
\end{itemize}
Across all scenarios, the robot must navigate through the context-dependent crowds to reach a designated goal.\\[0.1cm]
\noindent{\textbf{Baselines:} standard ROS~1 \texttt{move\_base} with shortest-path $\text{A}^*$ (base), and a socially-aware variant integrating human proxemic costs into path planning~\cite{lu2014layered} (social), both using TIAGo's default TEB local planner (max speed $0.5$~m/s).\\[0.1cm]
%
%\noindent\textbf{Causal Planner:} We set $(\lambda_\delta, \lambda_D, \lambda_L) = (1, 10, 5)$~\cite{castri2026causality}, to weight Euclidean distance ($\lambda_\delta$), crowd density ($\lambda_D$), and battery loss ($\lambda_L$), prioritizing proactive crowd avoidance as detailed in Sec.~\ref{sec:causal-heuristic}.\\[0.1cm]
%
\noindent\textbf{Hypotheses:} 
\begin{itemize}
    \item Effectiveness: The system yields safer paths and higher task success rates than the baselines across both simulated and physical environments.
    \item Adaptability \& Generalization: The invariant causal structure transfers effectively to new topological layouts with unseen human activities (sim-to-sim) and physical environments (sim-to-real) through lightweight parameter adaptation, avoiding structural discovery from scratch.
    \item Efficiency: The system is efficient, utilising a light-data parameter update process for rapid real-world adaptation and performing causal inference fast enough for real-time onboard execution.
\end{itemize}
\noindent\textbf{Metrics:} Task Success (binary), Safety (human-robot proxemics compliance), Inference Time.\\[0.1cm]
\noindent\textbf{Hardware:} Lenovo Legion (i7-10875H, RTX 2060, 16\,GB RAM), Ubuntu 20.04, ROS Noetic.

\begin{figure*}[t]
    \centering
    \begin{subfigure}{.687\textwidth}
    \includegraphics[trim={0cm 3cm 0cm 4cm}, clip, width=\columnwidth]{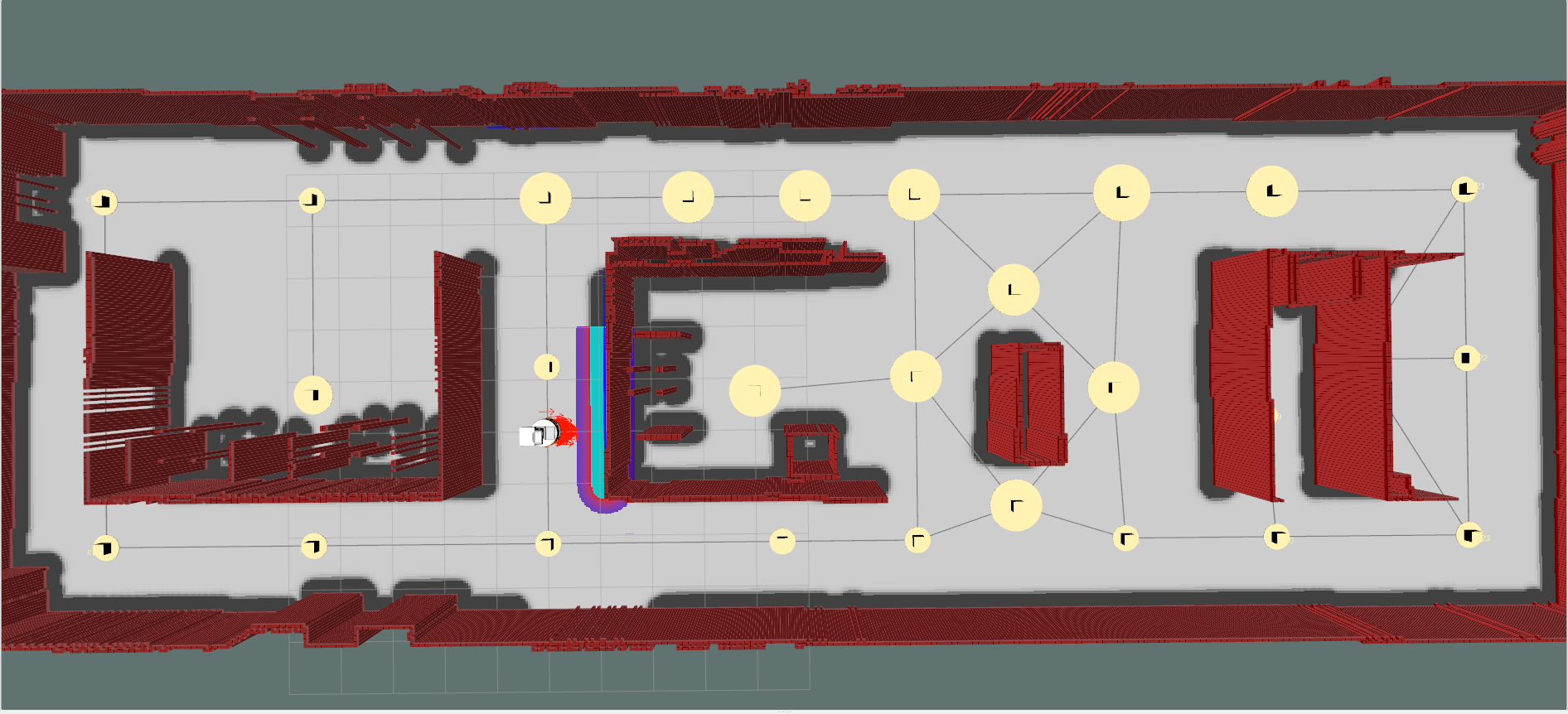}\\[0.1cm]
    \includegraphics[trim={0cm 0cm 0cm 0cm}, clip, width=\columnwidth]{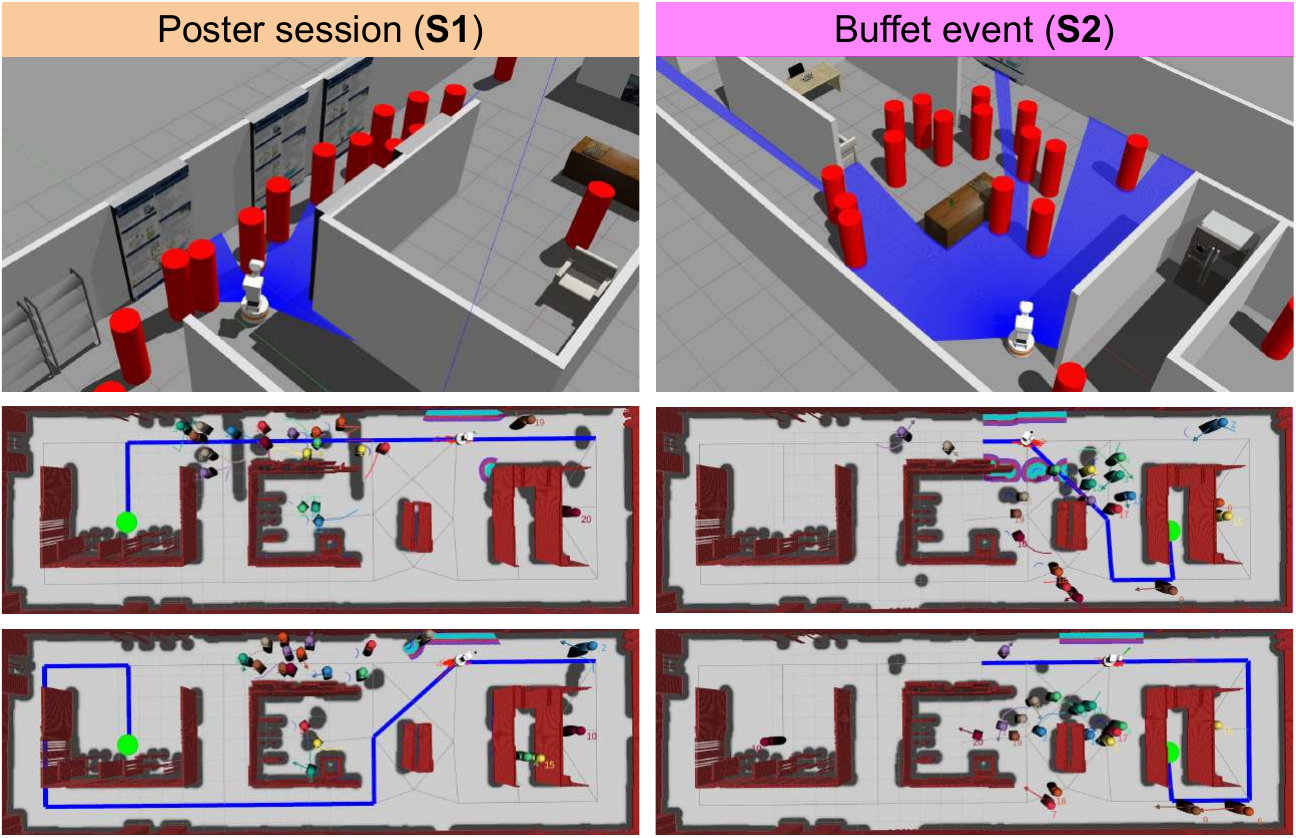}
    \caption{\textbf{Conference}}
    \label{fig:simulator-examples-conference}
    \end{subfigure}
    \hfill % Adds space between the two main columns
    \begin{subfigure}{.3055\textwidth}
    \includegraphics[trim={0cm 0.1cm 9.7cm 0cm}, clip, width=\columnwidth]{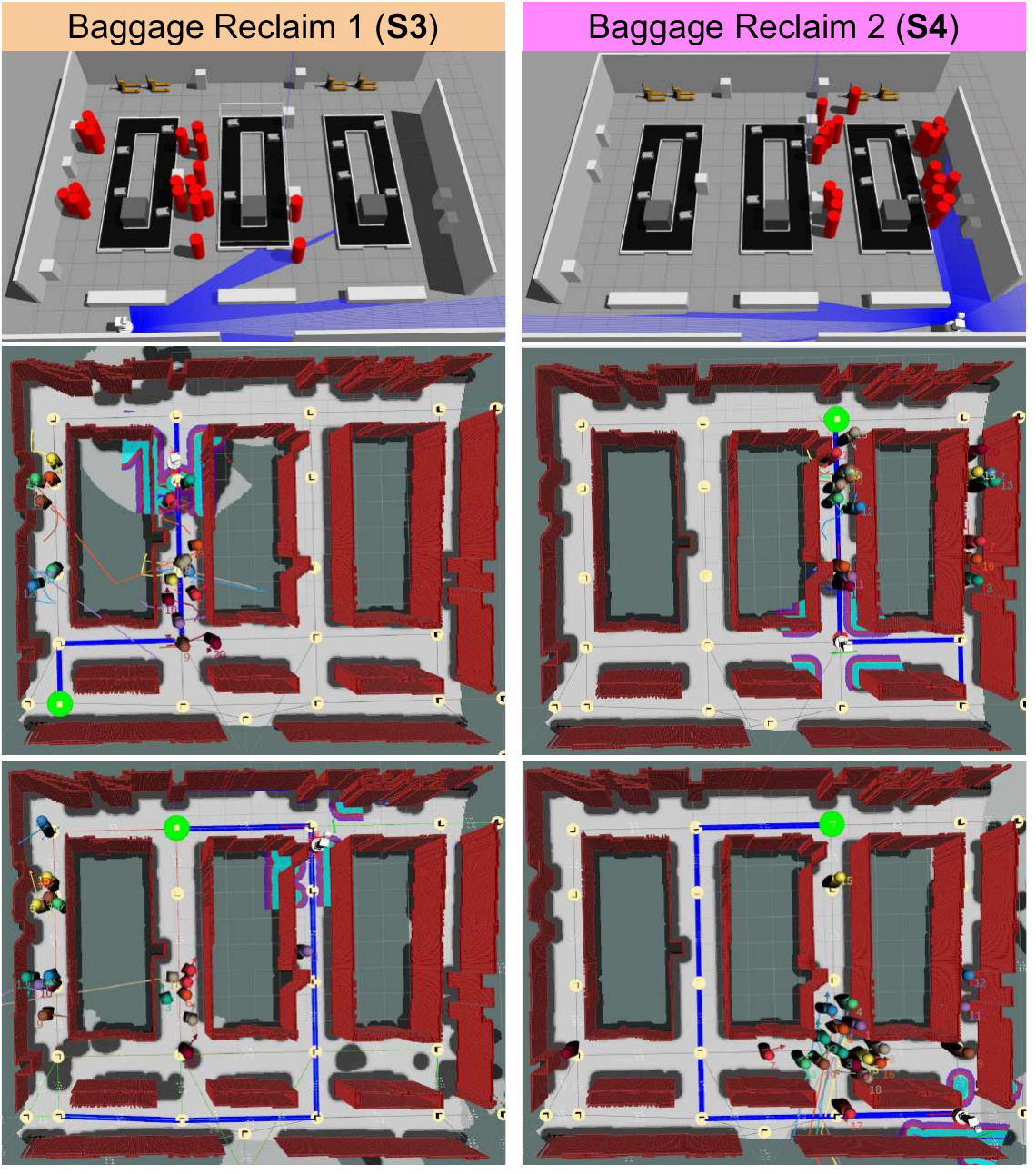}
    \caption{\textbf{Airport}}
    \label{fig:simulator-examples-airport}
    \end{subfigure}
    \caption{Simulated scenarios. (a) Conference: [Row 1] Topological map with navigation waypoints (yellow nodes). [Row 2] Snapshots of S1 (poster session) and S2 (buffet event). [Row 3] Baseline shortest path. [Row 4] Causal proactive planning. (b) Airport: [Row 1] Snapshot of S3 (baggage reclaim). [Row 2] Baseline shortest path. [Row 3] Causal clear route.}
    \label{fig:simulator-examples}
\end{figure*}

\subsection{Learning Scenarios}\label{sec:exp-learning}
Based on simulated conference-like environments (S1 and S2), we learn a causal structure that captures the main physical and operational dependencies (i.e. edges as shown in Fig.~\ref{fig:exp-cm}~[left]). Context ($T$, $W$) governs human presence $D$, while density $D$ and obstacles $O$ force deceleration in speed $S$, which subsequently modulates battery consumption $L$. Following causal invariance principles~\cite{scholkopf2021toward, rahimi2025sim}, these high-level mechanisms remain stable across environments, whereas environment-specific variations are captured by updating the CPD parameters. This structural invariance assumption holds as long as underlying dynamics and variables remain consistent, whereas unmodeled physical factors (e.g., battery temperature) or qualitative behavioral shifts (e.g., adversarial human obstruction) would require online causal structure re-learning. By freezing this underlying structure, sim-to-real transfer only requires a rapid parameter update, avoiding data-hungry causal discovery in the physical world.
%we avoid data-hungry causal discovery  in the physical world}, reducing the sim-to-real adaptation process to a rapid parameter update.}

\subsection{Data Generation and Causal Learning}\label{sec:exp-discovery}
For training data, we collected one hour of simulated data per scenario, during which the robot navigated the simulated environment using a standard A\textsuperscript{*} baseline planner. We recorded the robot's internal state ($S$, $L$), and contextual data ($T$, $W$, $O$). The raw data was post-processed to derive people density for each waypoint $W$ as $D = (\text{People in } W) / (\pi r_{W}^{2})$, where $r_W$ is the waypoint radius scaled to the local navigable space (e.g., narrower for doors, wider for corridors).
\begin{figure}
\centering
\begin{subfigure}{0.5\columnwidth}
    \includegraphics[trim={0cm 0.15cm 0cm 0cm}, clip, width=\columnwidth]{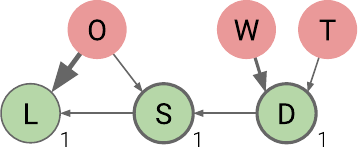}
\end{subfigure}
\begin{subfigure}{.2\columnwidth}
\vspace{1.5mm}
    \includegraphics[trim={23cm 4cm 23cm 9cm}, clip, width=\columnwidth]{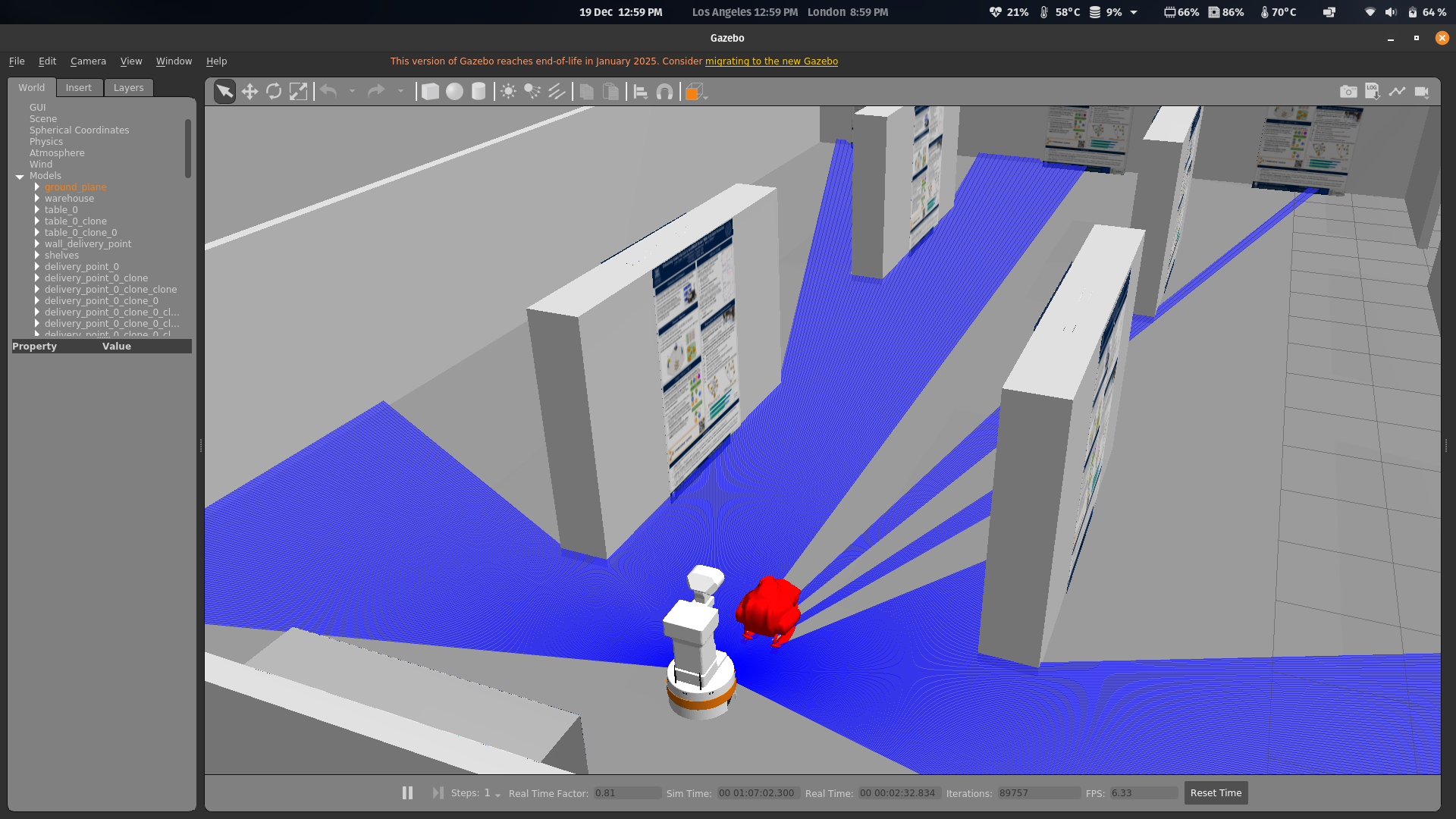}
\end{subfigure}
\caption{[Left] Retrieved causal model with context (red) and system (green) variables; edge thickness/borders mark dependency strength, and numbers show time lags ($0$ if unlabelled). [Right] TIAGo avoiding an unexpected obstacle.}
\label{fig:exp-cm}
\end{figure}
From CausalFlow\footref{foot:causalflow}, we used J-PCMCI\textsuperscript{+}~\cite{pmlr-v216-gunther23a} to learn the causal model using the combined data from both scenarios. This method was chosen for its ability to correctly model contextual variables as external drivers ($T, W, O$) that influence the system ($S$, $L$, $D$) without being influenced by it. Assuming a first-order Markov process---consistent with human spatial modelling literature~\cite{asahara2011pedestrian}---we ran conditional independence tests using RegressionCI\footnote{{\fontsize{6.5pt}{7.5pt}\selectfont Tigramite: \url{https://github.com/jakobrunge/tigramite}}} with a standard significance threshold ($\alpha=0.05$). The resulting graph (Fig.~\ref{fig:exp-cm}, left) matched our expected dependencies (Sec.~\ref{sec:exp-learning}).

Finally, continuous variables were discretised for parameter learning via MLE using the pyAgrum\footnote{{\fontsize{6.5pt}{7.5pt}\selectfont pyAgrum: \url{https://pyagrum.gitlab.io/}}} library.

\subsection{Causality-Driven Path Planning}\label{sec:causal-heuristic}
To route the robot proactively, the expectations inferred from the causal model (Fig.~\ref{fig:exp-cm}~[left]) are embedded into the A$^*$ global planner as edge costs. The cost for traversing an edge $(u, v)$ between two consecutive waypoints is defined as:
\begin{equation}\label{eq:causal-inference-cost}\footnotesize
    c(u, v) = \lambda_\delta \cdot \delta(u, v) + \lambda_D \cdot \hat{D}(v) + \lambda_L \cdot \hat{L}(u, v)
\end{equation}
where $\delta(u, v)$ is the Euclidean distance between $u$ and $v$, $\hat{D}(v) = \mathbb{E}[D \mid \text{do}(T = t), W = v]$ is the expected human density at waypoint $v$ at time $t$, and $\hat{L}(u, v) = \mathbb{E}[L \mid \text{do}(S = s_{\max})]$ is the expected battery drain across $(u, v)$ at maximum cruising speed $s_{\max} = 0.5~\text{m/s}$. The coefficients $\lambda_\delta$, $\lambda_D$, and $\lambda_L$ balance path length, crowd avoidance, and energy consumption. The search is guided by the heuristic $h(u) = \lambda_\delta \cdot \delta(u, g)$, i.e., the weighted Euclidean distance from $u$ to the goal $g$. Because $\hat{D}$ and $\hat{L}$ are non-negative and $\delta$ satisfies the triangle inequality, $h$ is admissible and consistent; $\text{A}^*$ therefore returns the path of minimum expected cost. Note that this optimality holds with respect to the forecasted costs, not the costs actually incurred during execution. Path planning is evaluated at the start of each navigation run. We set $(\lambda_\delta, \lambda_D, \lambda_L) = (1, 10, 5)$~\cite{castri2026causality}, to weight Euclidean distance ($\lambda_\delta$), crowd density ($\lambda_D$), and battery loss ($\lambda_L$), prioritizing proactive crowd avoidance as detailed.

\section{SIMULATION RESULTS}\label{sec:exp-sim}
\begin{figure*}[t]
    \centering
    \begin{subfigure}{0.49\textwidth}
        \centering
        \includegraphics[trim={0cm 0cm 0cm, 0cm},clip,width=\textwidth]{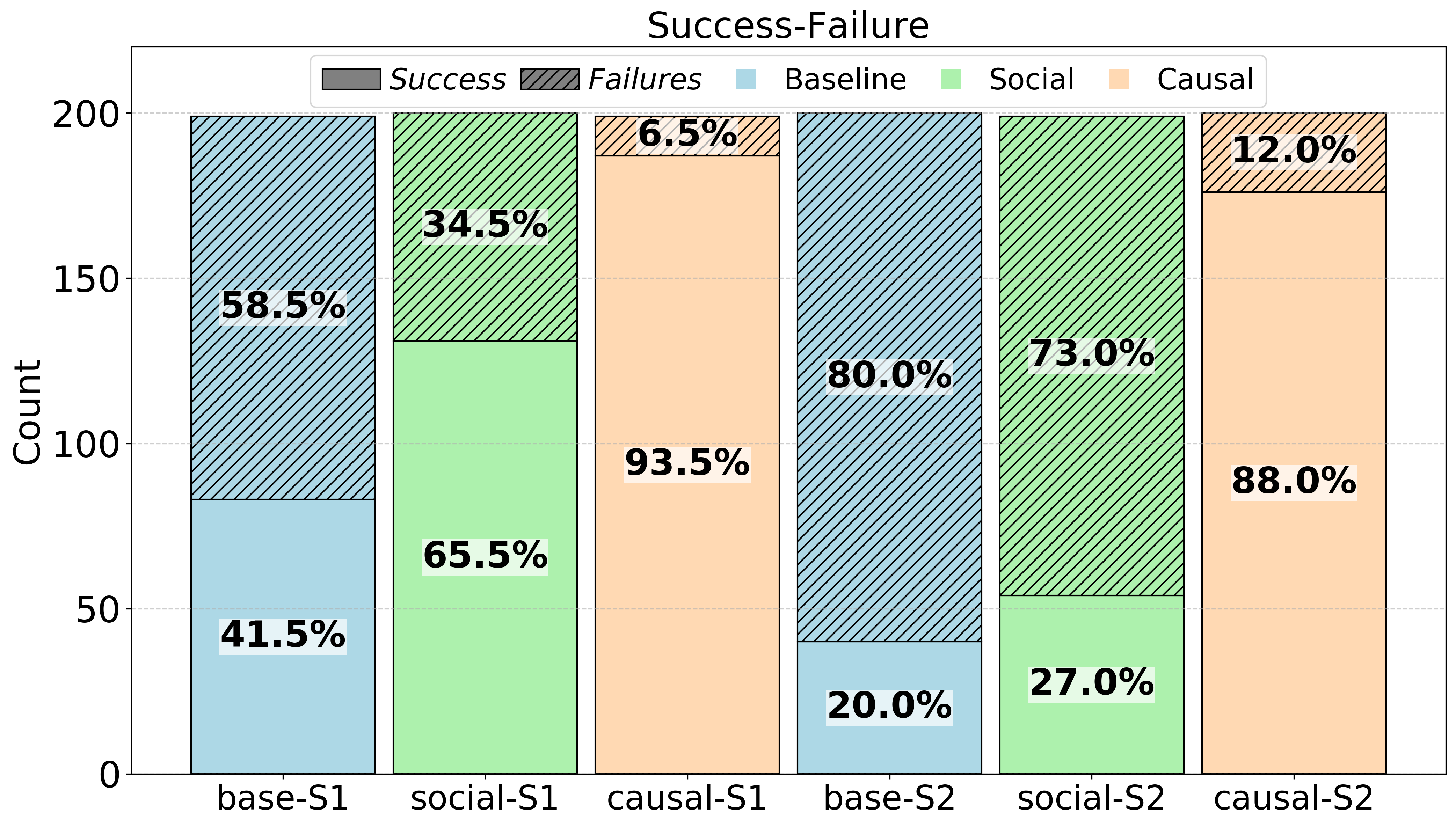}\\[0.05cm]
        \includegraphics[trim={2.3cm 0cm 2.3cm, 0cm},clip,width=\textwidth]{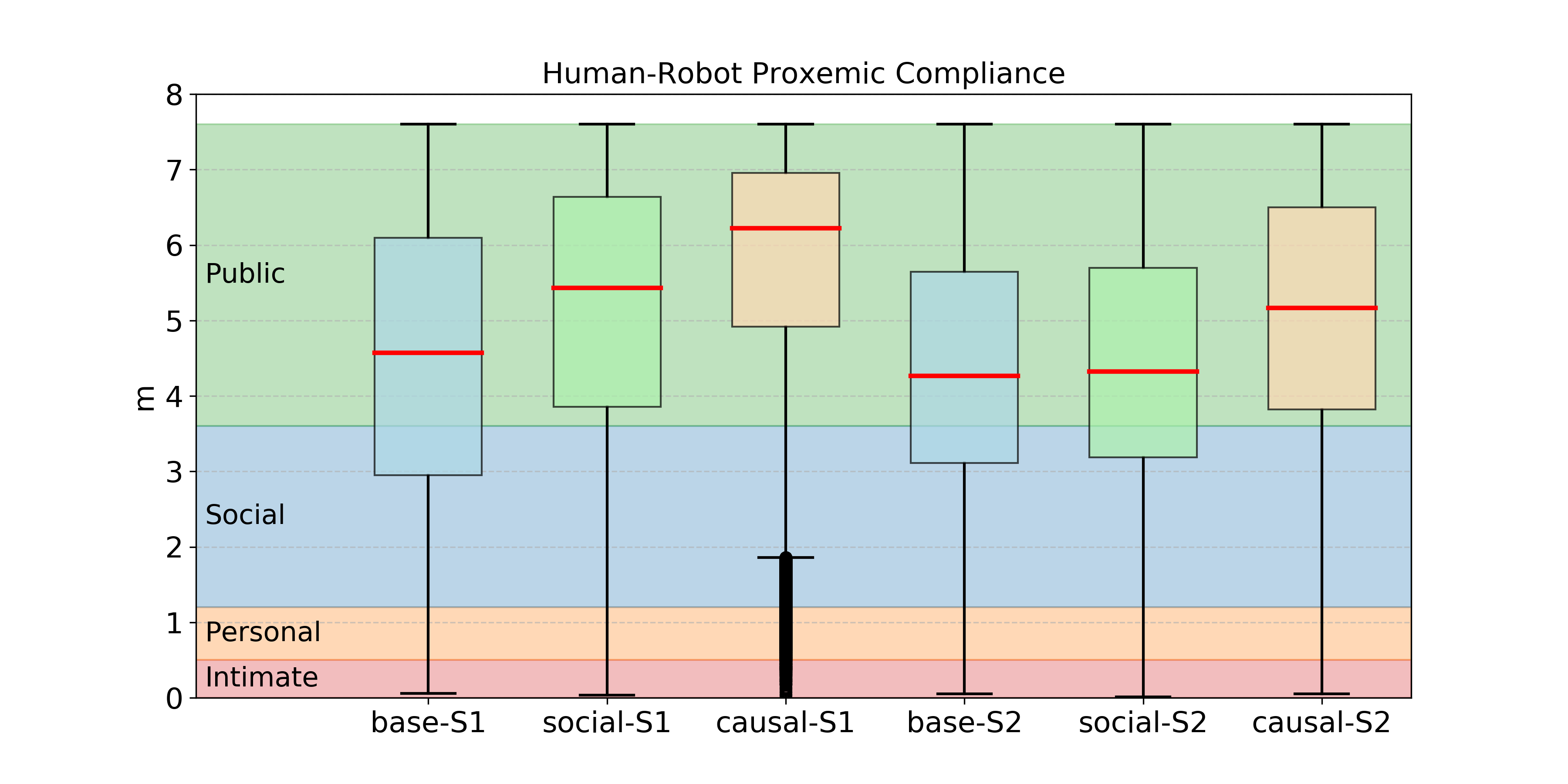}
        \caption{\textbf{Conference}}
        \label{fig:exp-sim-results-conference}
    \end{subfigure}
    \hfill % Adds space between the two main columns
    \begin{subfigure}{0.49\textwidth}
        \centering
        \includegraphics[trim={0cm 0cm 0cm, 0cm},clip,width=\textwidth]{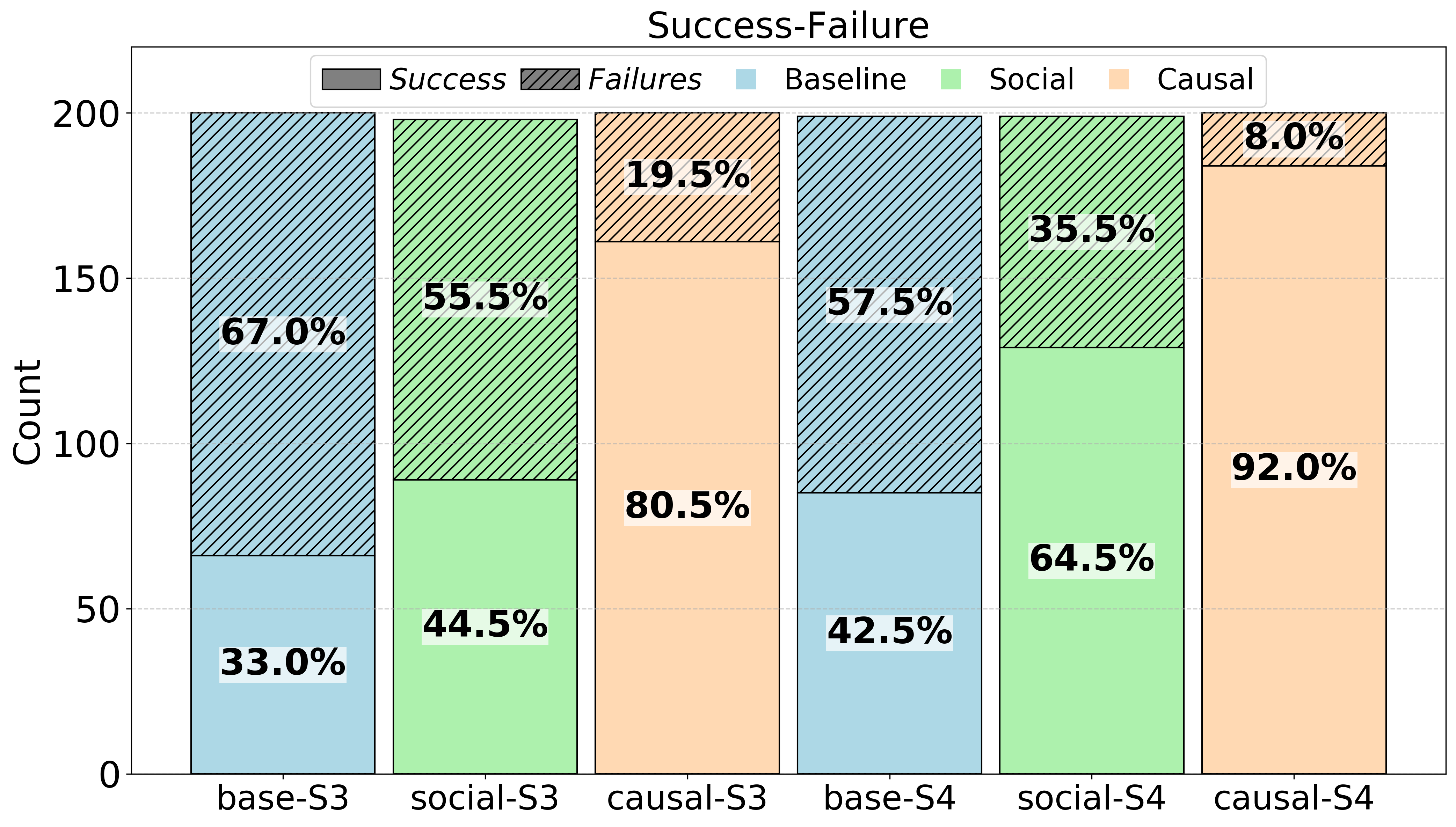}\\[0.05cm]
        \includegraphics[trim={2.3cm 0cm 2.3cm, 0cm},clip,width=\textwidth]{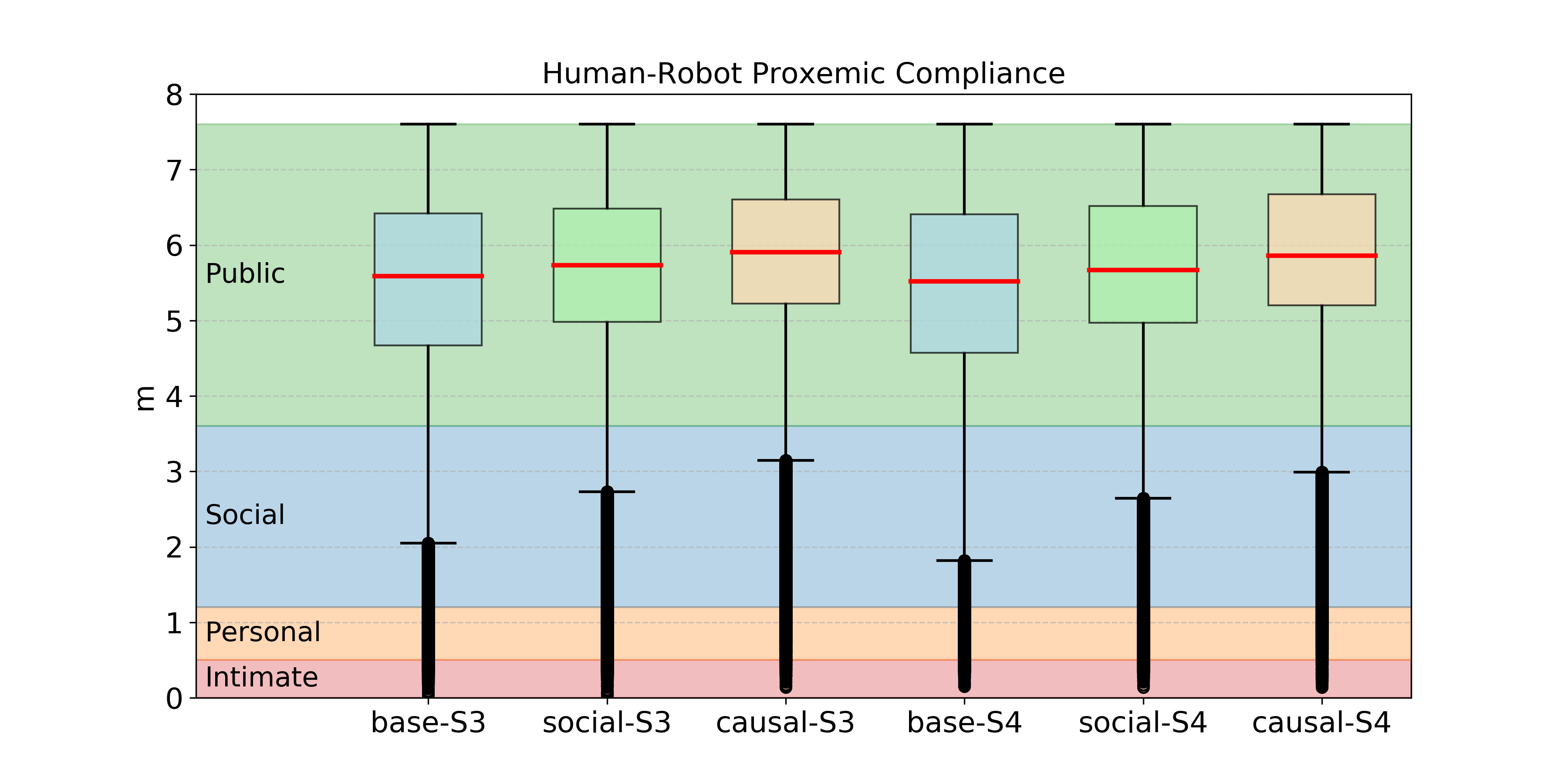}
        \caption{\textbf{Airport}}
        \label{fig:exp-sim-results-airport}
    \end{subfigure}
    %\caption{Simulation navigation performance in the \textbf{Conference}~(a) and \textbf{Airport}~(b) environments. (Top) Success/failure distribution. (Bottom) Human-robot proxemic compliance. Background colours representing Hall's proxemic zones~\cite{hall1968proxemics}: Intimate (red), Personal (orange), Social (blue), and Public (green).}\label{fig:exp-sim-results}
    \caption{Navigation performance across $200$ trials per method in the simulated Conference~(a) and Airport~(b) environments. (Top) Success and failure distributions for base, social, and causal planners. (Bottom) Human-robot proxemic compliance distributions across Hall's proxemic zones~\cite{hall1968proxemics}: Intimate (red), Personal (orange), Social (blue), and Public (green).}
\end{figure*}
This section presents the first stage of our experimental evaluation, i.e., the simulation results. Specifically, we validate the navigation performance of the proposed causality-enhanced framework and demonstrate the causal model outperforms a standard Bayesian network for inference tasks. 
To assess the effectiveness of our path planning and task execution, we evaluated the robot's navigation behaviour using our Gazebo-based social simulator (Fig.~\ref{fig:approach}, left). A topological map (Fig.~\ref{fig:simulator-examples}~top) was employed to facilitate both TIAGo and the simulated humans spatial navigation. To test the sim-to-sim transfer capabilities of our approach, we evaluate whether the invariant causal graph structure learned from the conference scenarios (S1 and S2) can be directly applied to a completely new, unseen simulated airport environment (S3 and S4). Specifically, the airport model is adapted using a 2-minute observational data window to recalibrate the tabular CPD parameters over the invariant structure, mirroring the real-world adaptation protocol.
The robot acts as a service agent tasked with completing a series of waypoint-to-waypoint navigation tasks. For each task, the robot must travel from a designated start node to a goal node across the topological graph. To emulate the unpredictability of physical human environments, each edge connecting two waypoints carries a predefined probability of static obstacle appearance $P(O)$---simulating items such as backpacks temporarily left in corridors by conference attendees (Fig.~\ref{fig:exp-cm} right). Unexpected obstacles forces the robot to execute sudden avoidance manoeuvres, which directly alters its speed profile and increases its overall battery consumption.

As shown in Fig.~\ref{fig:simulator-examples}, the system correctly anticipated context-driven congestion trends during the poster session (S1) and buffet event (S2), as well as around the baggage reclaim belts (S3). The causal planner (bottom row) proactively inferred these high-density zones and selected detours to avoid them, maintaining smooth and safe navigation. In contrast, the base planner (middle row), being blind to the causal drivers of density, consistently attempted to traverse the congested areas.

\subsection{Navigation Performance}\label{sec:esp-sim-results}
We evaluated the framework against the base and social baselines across $200$ simulated navigation trials per scenario.
\subsubsection{Task Success} %The Success-Failure distribution (Fig.~\ref{fig:exp-sim-results} left) shows the causal planner significantly outperformed the two baselines. In S1, the task success improved from $41.5$\% to $93.5$\%. In the highly dynamic S2, where the baseline failed in $80$\% of the trials, the causal approach maintained an $88$\% success rate. This is further supported by the completion percentage analysis (Fig.~\ref{fig:exp-sim-results} centre), which demonstrates that the baseline on average fails midway through the tasks (with median path completion ratios of $0.55$ and $0.42$ for S1 and S2, respectively), while the causal approach consistently reaches the goal (median completion ratios of $1$).
For both the Conference (Fig.~\ref{fig:exp-sim-results-conference}, top) and Airport (Fig.~\ref{fig:exp-sim-results-airport}, top) domains, the causal planner consistently outperformed the baselines across all scenarios (S1, S2, S3, and S4). This confirms parameter recalibration across distinct topologies and crowd dynamics performing lightweight tabular CPD adaptation over the invariant structure.
\subsubsection{Proxemic Compliance} %The human-robot proxemic compliance evaluation (Fig.~\ref{fig:exp-sim-results} right) reveals that the baseline's interquartile range (IQR) extends down into the undesired ``Social'' space ($1.2$--$3.6$~m) in both scenarios. Conversely, the causal planner by proactively anticipating congestion and selecting detours, shifts the entire IQR of human-robot distances safely into the ``Public'' zone ($>3.6$~m).
In both environments (Figs.~\ref{fig:exp-sim-results-conference}--\ref{fig:exp-sim-results-airport}, bottom), while all methods operate mainly within the ``Public'' zone ($> 3.6$~m), the causal planner minimizes personal and social space intrusions by proactively taking detours.

\subsection{Causal vs. Bayesian Prediction}
A core advantage of our framework is its ability to disentangle true causation from spurious correlation. We demonstrate this via a confounded inference task: predicting battery consumption ($L$). As shown in Fig.~\ref{fig:exp-cm} (left), the selected speed~$S$ directly affects~$L$, but this relationship is confounded by unexpected obstacles~$O$.

Standard Bayesian inference would simply estimate the depletion probability based on observed speed as follows:
\begin{equation}\label{eq:bn-query}
P(L \mid S) = \sum P(L \mid S, O)P(O \mid S).
\end{equation}
This approach though implicitly introduces some bias due to confounding obstacle $O$. Our framework instead uses the \mbox{{\em do-operator}}~\cite{pearl2009causality} to predict the causal effect of a hypothetical speed on the battery consumption, regardless of any obstacle:
\begin{equation}\label{eq:cm-query}
P(L \mid  \text{do}(S)) = \sum P(L \mid S, O)P(O),
\end{equation}
which adjusts for the confounder $O$ by marginalising over its prior distribution rather than its conditional one.

To evaluate these approaches, we ran a series of simulated experiments with our Gazebo-based social simulator, where the robot performed a navigation task at a maximum speed \mbox{$S_\text{MAX} = 0.5$~m/s}, while varying the probability of encountering unexpected obstacles $P(O)$ from $10\%$ to $50\%$ in steps of $10\%$. For each obstacle probability, $50$ navigation tasks were executed to measure empirically the ground truth $L_\text{GT}$. We then queried the Bayesian model, $L_{\text{BN}} = \mathbb{E}[L \mid S_{\text{MAX}}]$, and the causal one, $L_{\text{CM}} = \mathbb{E}[L \mid \text{do}(S_{\text{MAX}})]$, using (\ref{eq:bn-query})~and~(\ref{eq:cm-query}).

Fig.~\ref{fig:exp-sim-battery} shows the causal prediction consistently yielded a lower Mean Absolute Error (MAE) across all $P(O)$ values. %Fig.~\ref{fig:exp-sim-battery} (bottom) also shows the causal model yielded more accurate predictions than the Bayesian network across the $50$ tasks for each $P(O)$ setting, outperforming the baseline in $80\%$ of the $250$ total runs. 
This result demonstrates that the causal model is more robust within confounded environments.

\begin{figure}
    \vspace{1.5mm}
    \centering
    %\begin{subfigure}{.795\columnwidth}
    \includegraphics[trim={0cm 0cm 0cm 0cm}, clip, width=0.5\columnwidth]{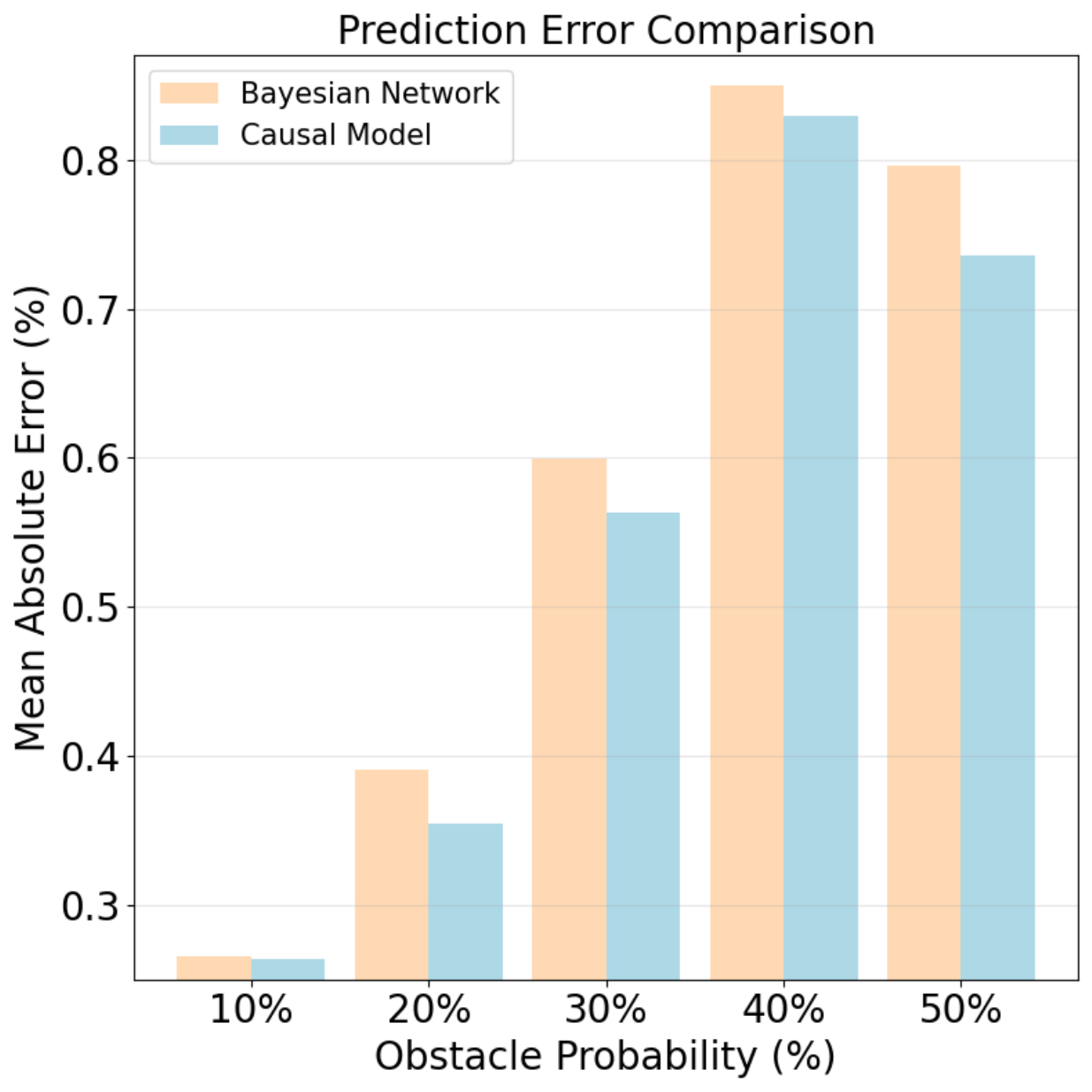}
    %\end{subfigure}\\
    %\begin{subfigure}{.795\columnwidth}
    %    \includegraphics[trim={0cm 14cm 0cm 0cm}, clip, width=\columnwidth]{resources/exp-sim-battery-count.pdf}
    %\end{subfigure}
    \caption{MAE of Bayesian and causal battery predictions for varying obstacle probabilities.} %(Bottom) Most accurate predictions count over 250 runs.}
    \label{fig:exp-sim-battery}
\end{figure}

%To show the practical impact of this bias, consider a $40$~m navigation task at $P_O = 30\%$, with an actual battery cost of $L_{GT} = 0.785$. The Bayesian prediction differed from this true cost by $\sim 8\%$. By adjusting for the confounder, the causal model reduced error to $1.6\%$---a $78.5\%$ relative reduction in error compared to the Bayesian network---confirming its superior reliability in confounded environments.
\section{REAL-WORLD RESULTS}\label{sec:exp-real}
\begin{figure*}[t]
\vspace{1.5mm}
    \centering
    % ================= COLUMN a: SETUP =================
    \begin{subfigure}{0.35\textwidth}
        \centering
        % 1. Map (Top)
        \includegraphics[trim={0cm 0cm 0cm 0.5cm}, clip, width=\linewidth]{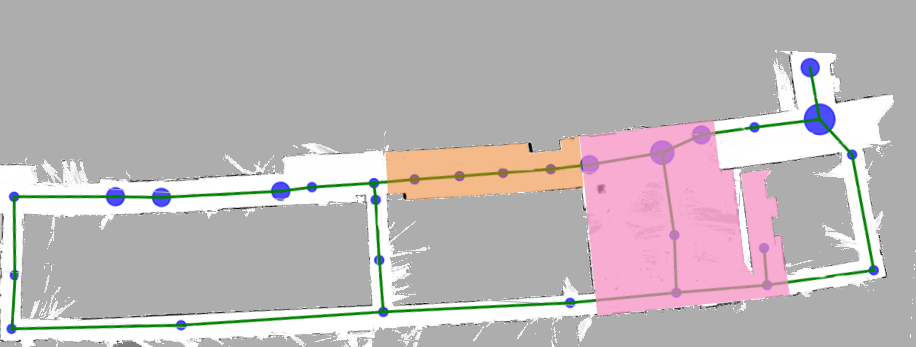}\\[0.1cm]
        % 2. Poster Setup Row (Middle)
        \begin{minipage}{0.73\linewidth}
            \includegraphics[trim={0cm 0cm 0cm 2cm}, clip, width=\linewidth]{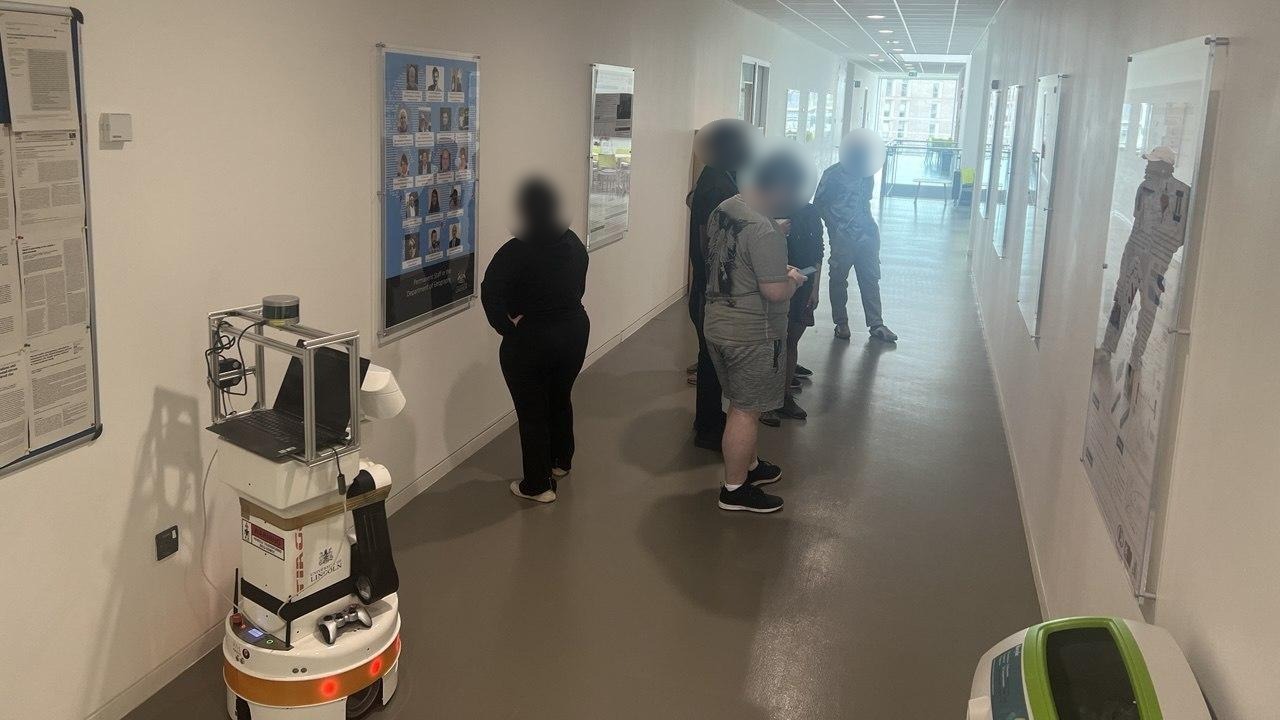}
        \end{minipage}%
        \hfill
        \begin{minipage}{0.26\linewidth}
            \includegraphics[trim={25cm 7.2cm 25cm 5cm}, clip, width=\linewidth]{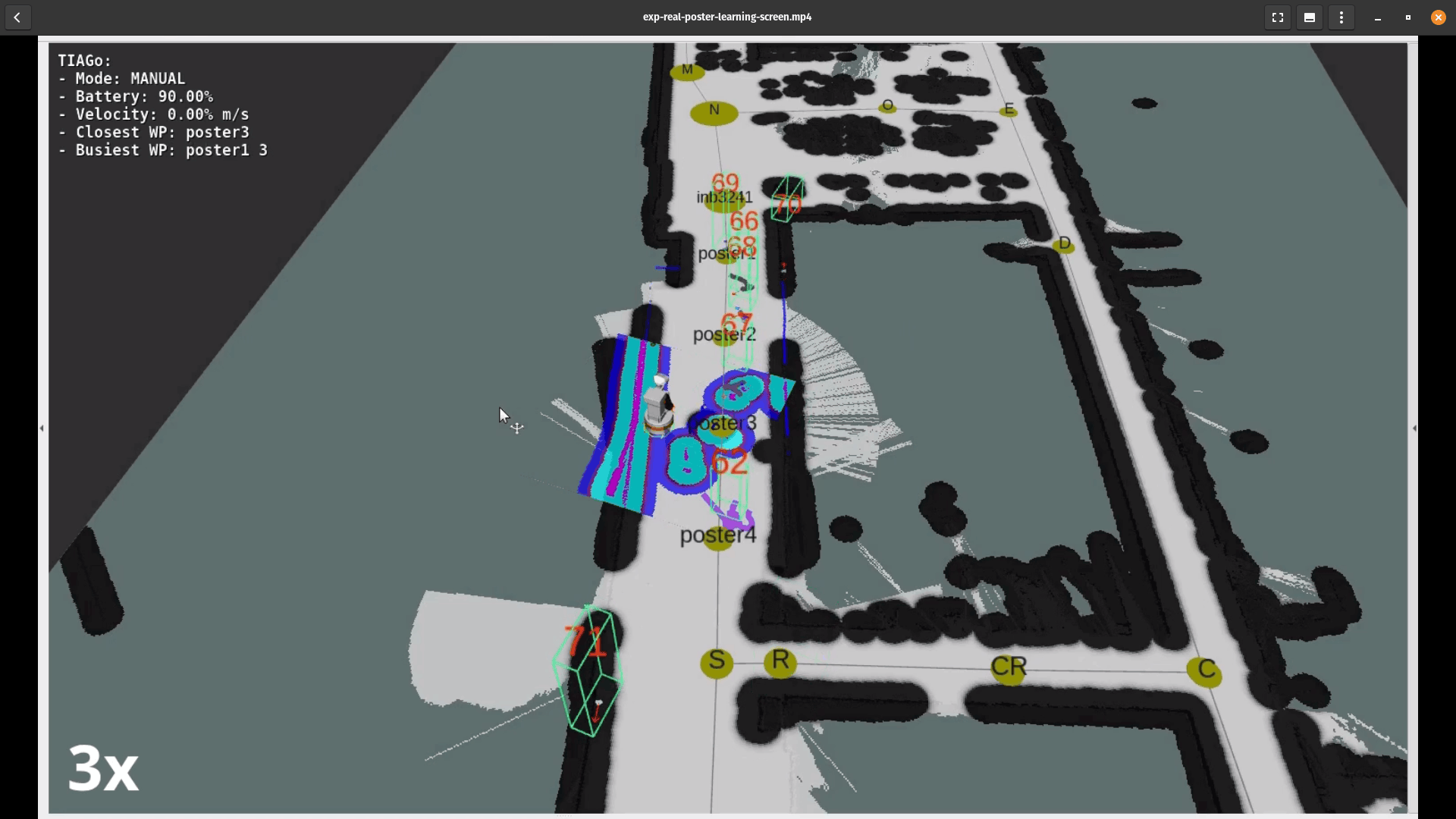}
        \end{minipage}
        \vspace{0.05cm}
        
        % 3. Buffet Setup Row (Bottom)
        \begin{minipage}{0.55\linewidth}
            \includegraphics[trim={0cm 0cm 0cm 0cm}, clip, width=\linewidth]{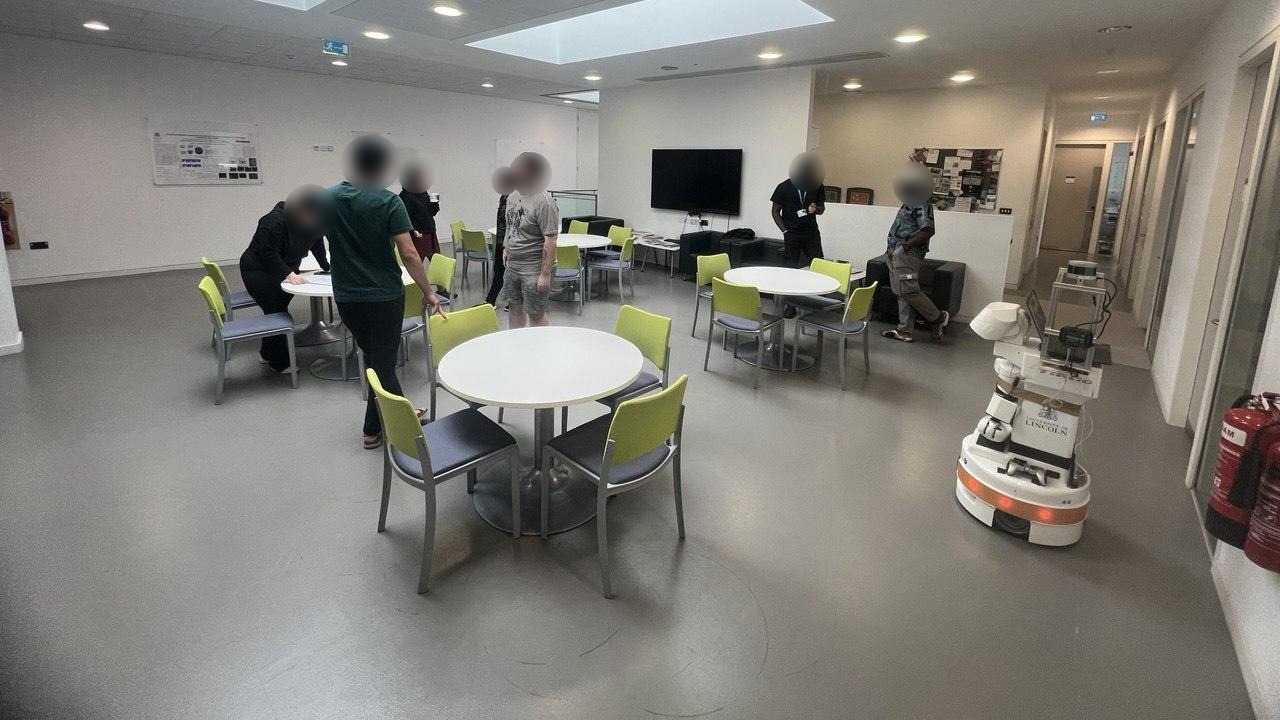}
        \end{minipage}%
        \hfill
        \begin{minipage}{0.43\linewidth}
            \includegraphics[trim={25cm 10cm 15cm 8cm}, clip, width=\linewidth]{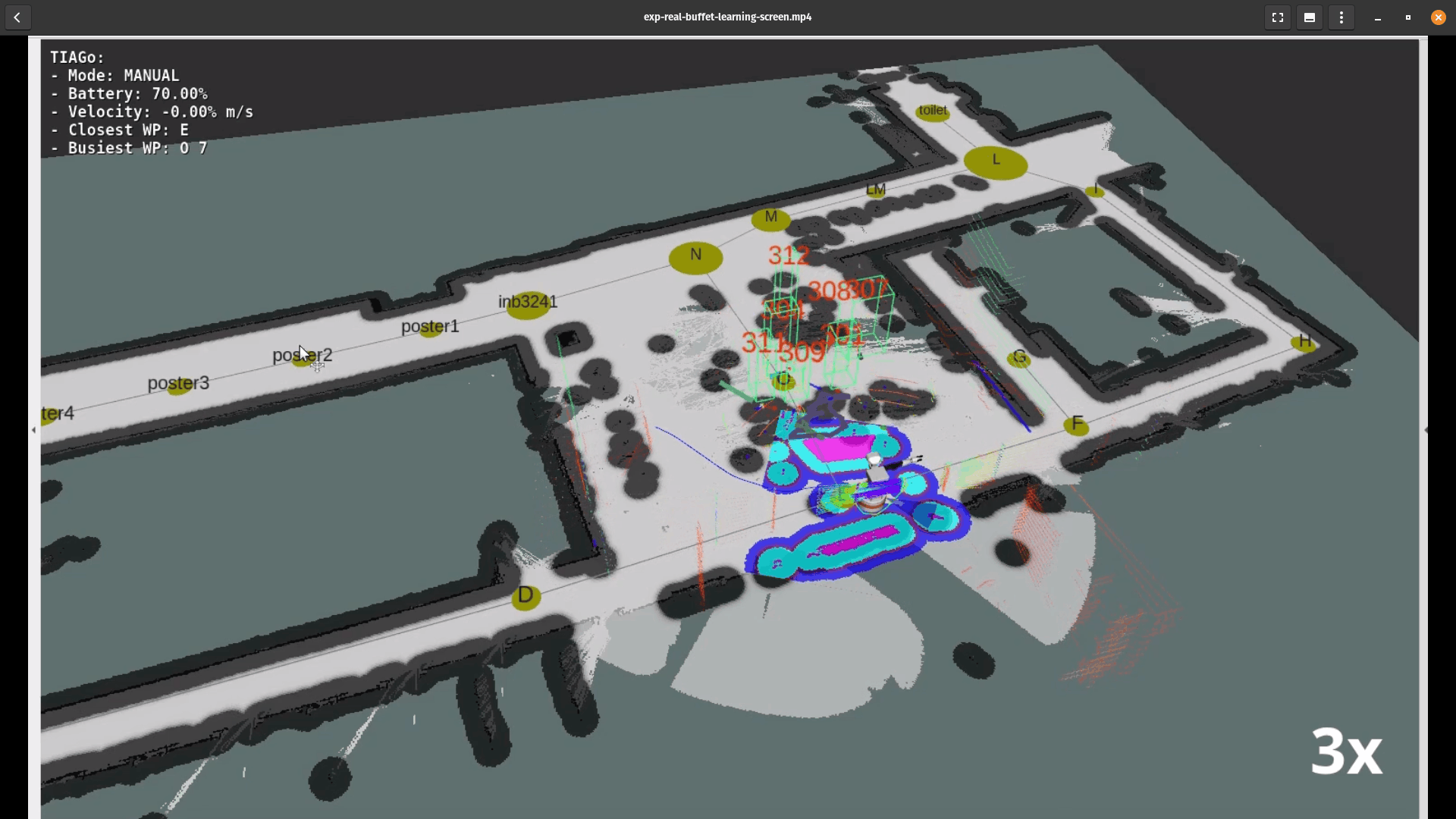}
        \end{minipage}
        
        \caption{Parameter update setup}
        \label{fig:exp-real-learning}
    \end{subfigure}
    \hfill
    % ================= COLUMN b: POSTER SCENARIO =================
    \begin{subfigure}{0.31\textwidth}
        \centering
        % 1. Map (Top)
        \includegraphics[trim={6cm 1.5cm 3cm 2cm}, clip, width=\linewidth]{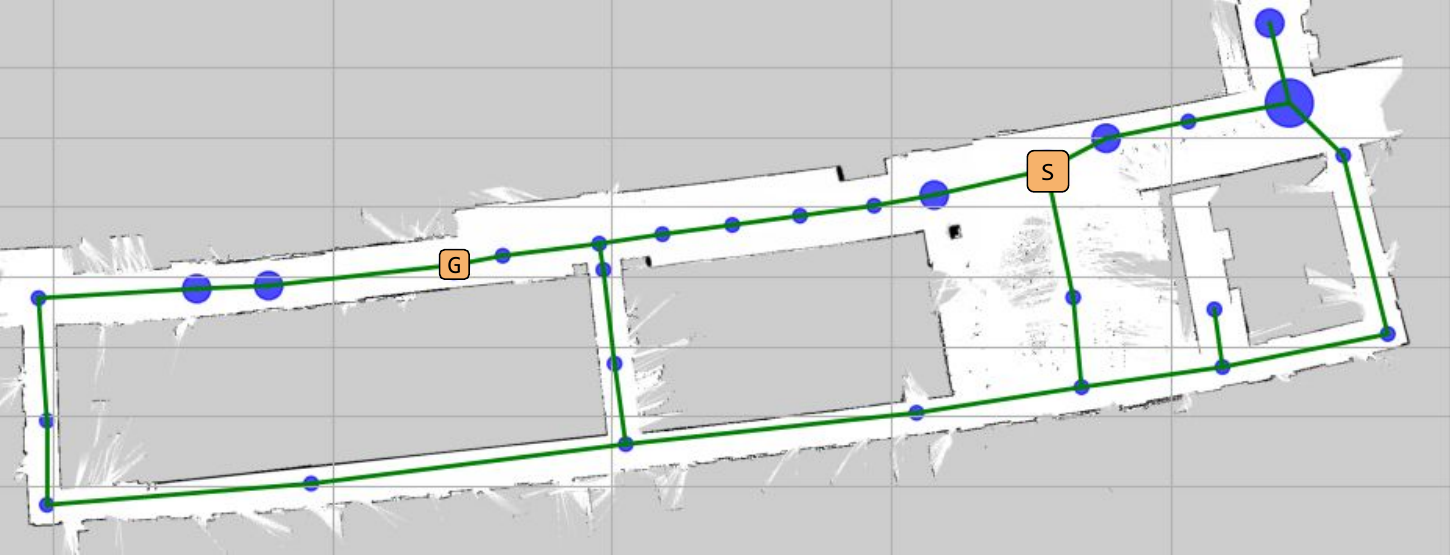}\\[0.1cm]
        % 2. Baseline Row (Middle)
        \begin{minipage}{0.76\linewidth}
            \includegraphics[trim={10cm 5cm 6cm 7.5cm}, clip, width=\linewidth]{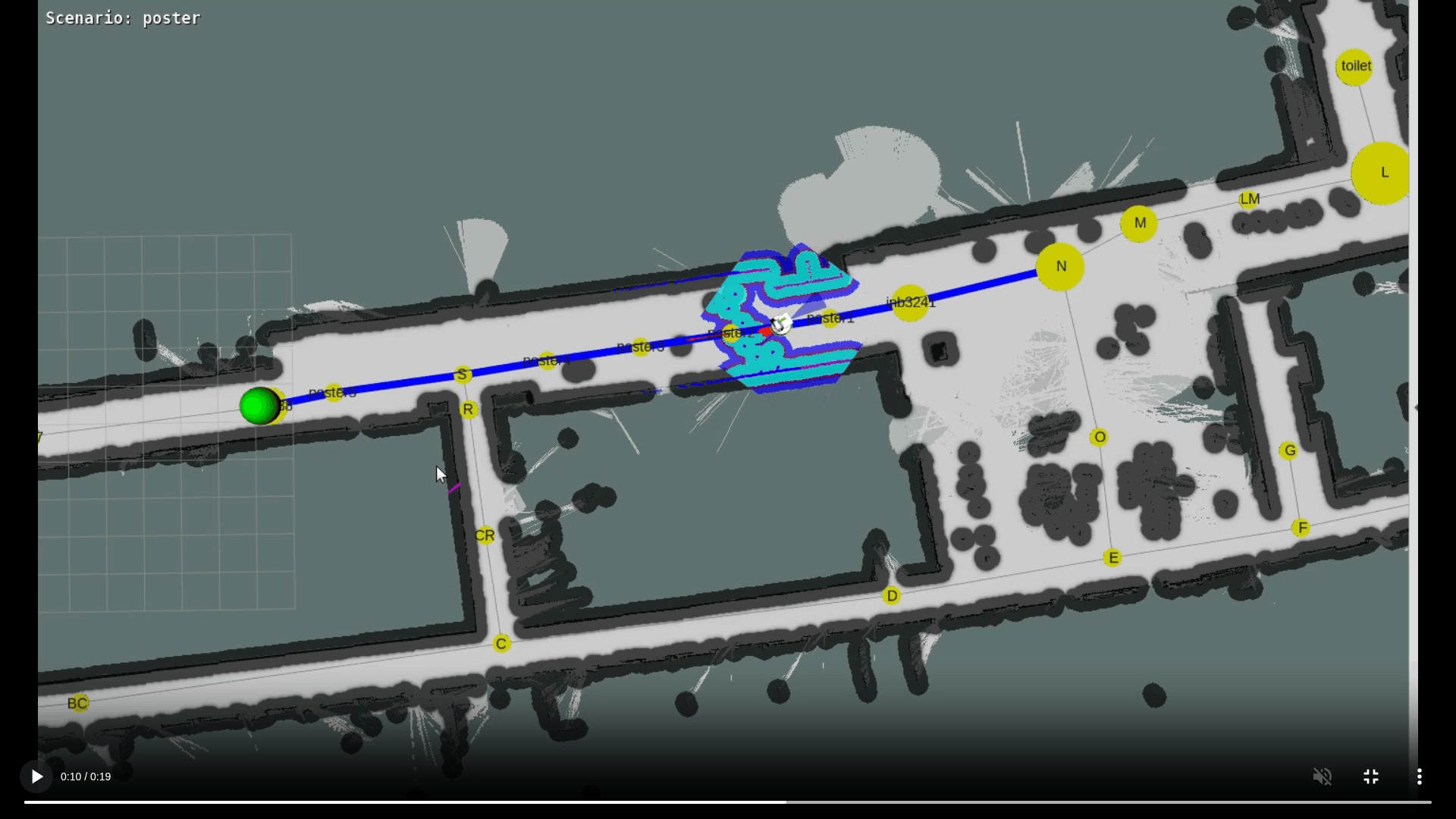}
        \end{minipage}%
        \hfill
        \begin{minipage}{0.23\linewidth}
            \includegraphics[trim={0cm 0cm 0cm 2.5cm}, clip, width=\linewidth]{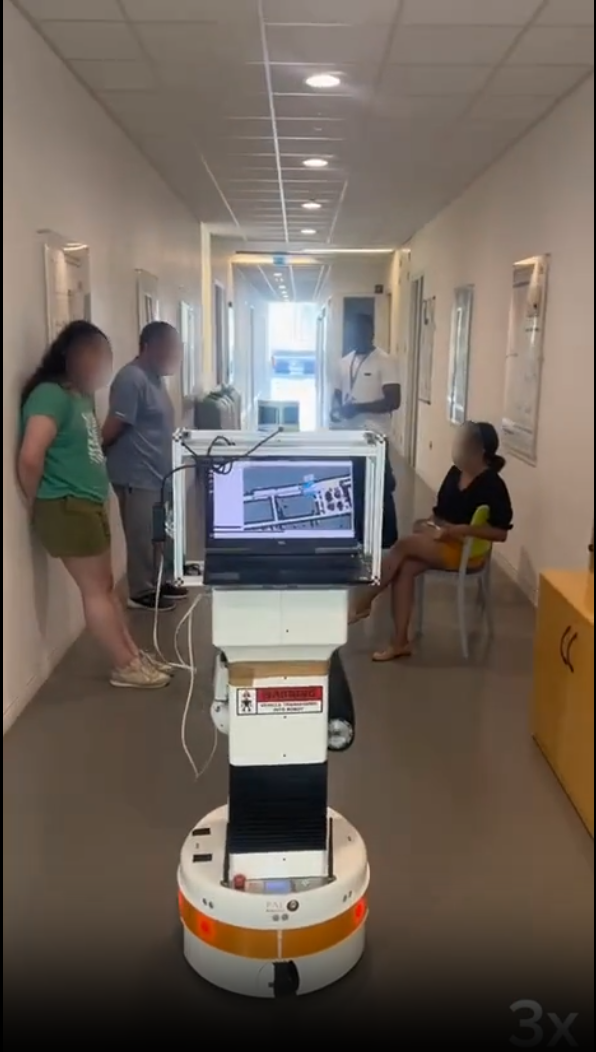}
        \end{minipage}
        \vspace{0.05cm}
        
        % 3. Causal Row (Bottom)
        \begin{minipage}{0.76\linewidth}
            \includegraphics[trim={10cm 5cm 6cm 7.5cm}, clip, width=\linewidth]{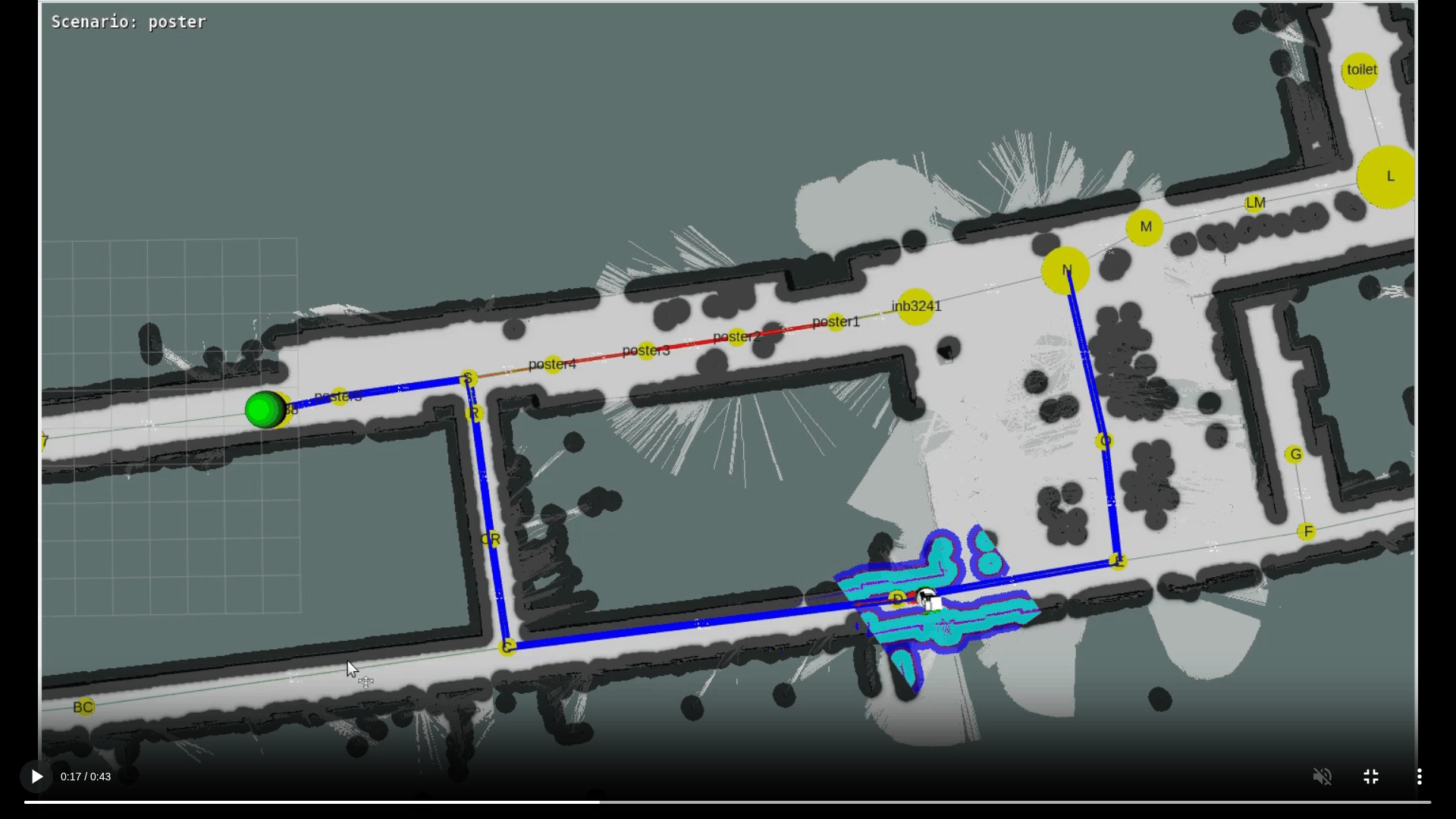}
        \end{minipage}%
        \hfill
        \begin{minipage}{0.23\linewidth}
            \includegraphics[trim={0cm 0cm 0cm 2.5cm}, clip, width=\linewidth]{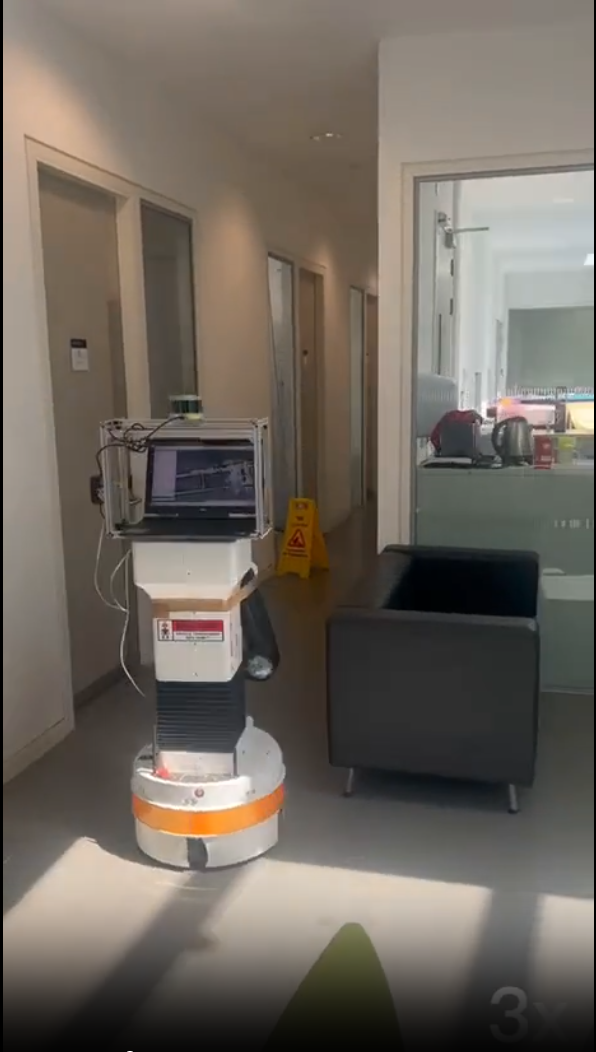}
        \end{minipage}
        
        \caption{S1---Poster Session} 
        \label{fig:exp-real-inference-poster}
    \end{subfigure}
    \hfill
    % ================= COLUMN c: BUFFET SCENARIO =================
    \begin{subfigure}{0.31\textwidth}
        \centering
        % 1. Map (Top)
        \includegraphics[trim={9cm 2.5cm 0cm 1cm}, clip, width=\linewidth]{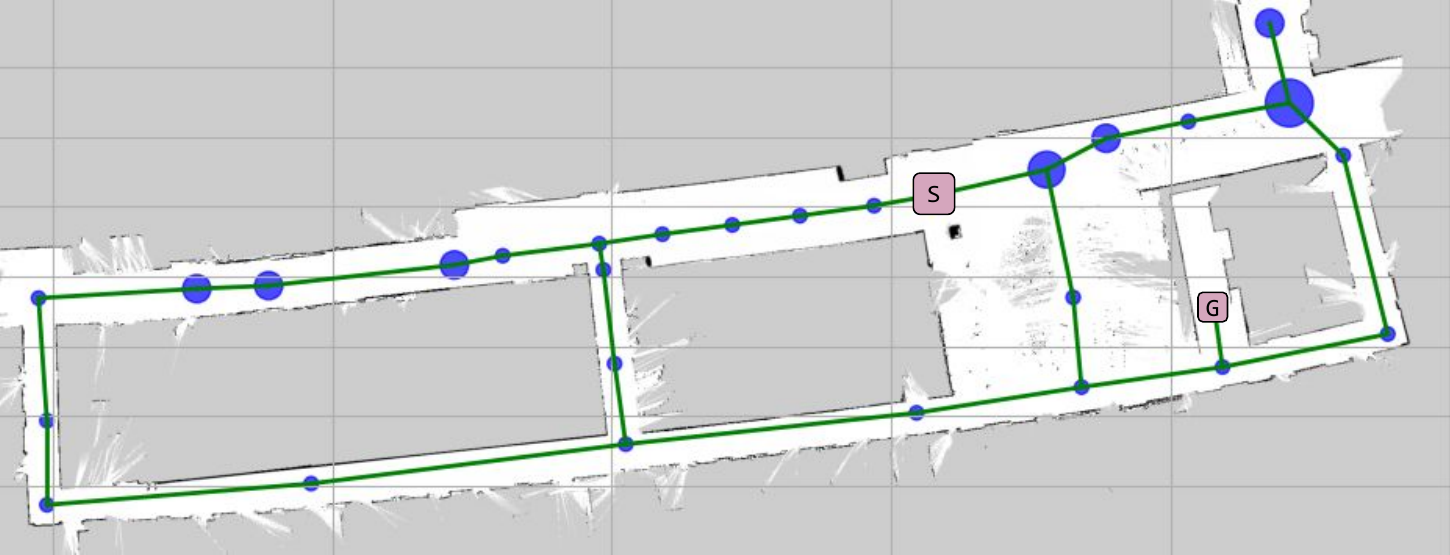}\\[0.1cm]
        % 2. Baseline Row (Middle)
        \begin{minipage}{0.76\linewidth}
            \includegraphics[trim={10cm 5cm 6cm 7.5cm}, clip, width=\linewidth]{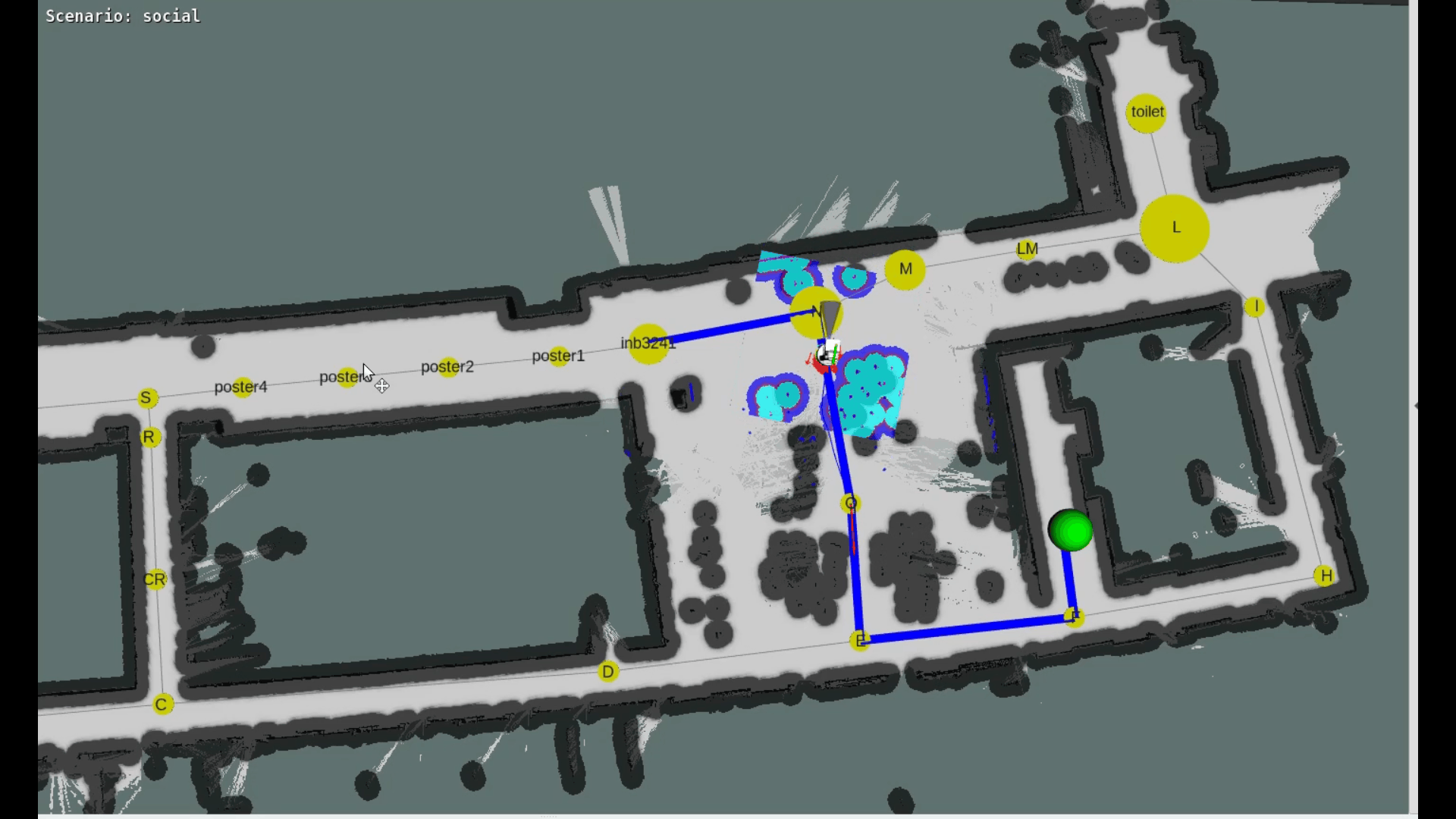}
        \end{minipage}%
        \hfill
        \begin{minipage}{0.23\linewidth}
            \includegraphics[trim={0cm 0cm 0cm 0cm}, clip, width=\linewidth]{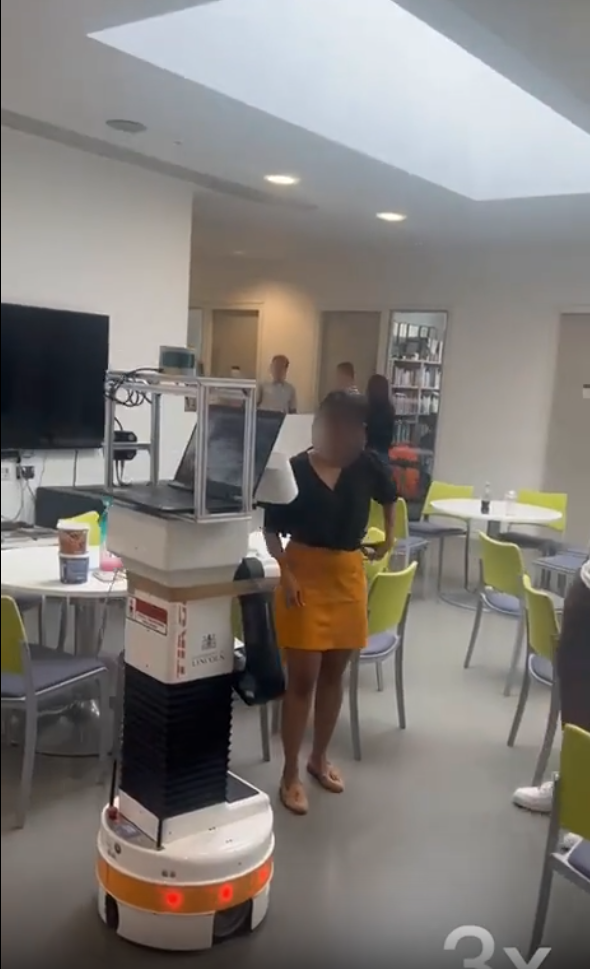}
        \end{minipage}
        \vspace{0.05cm}
        
        % 3. Causal Row (Bottom)
        \begin{minipage}{0.76\linewidth}
            \includegraphics[trim={10cm 5cm 6cm 7.5cm}, clip, width=\linewidth]{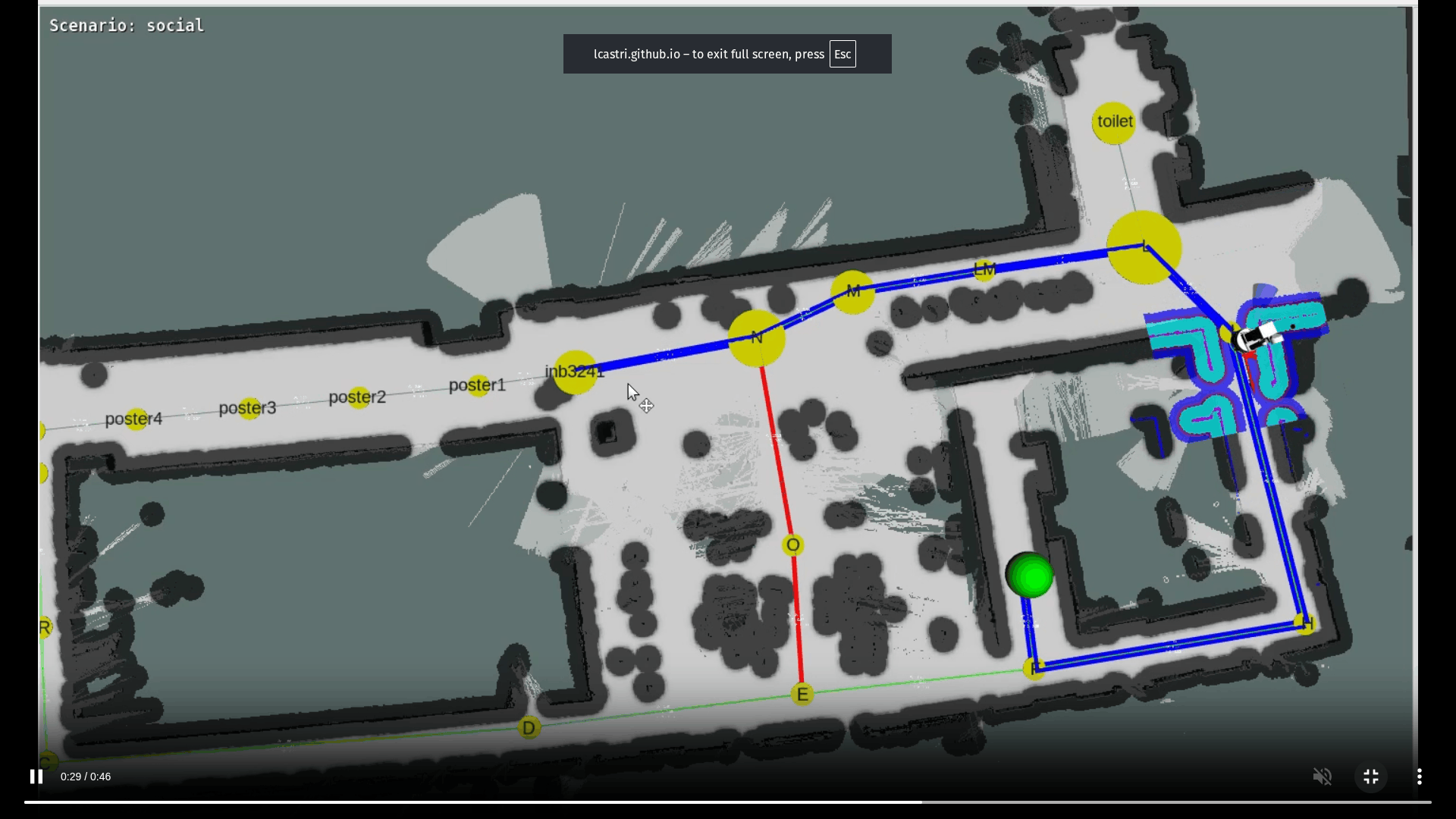}
        \end{minipage}%
        \hfill
        \begin{minipage}{0.23\linewidth}
            \includegraphics[trim={0cm 0cm 0cm 0cm}, clip, width=\linewidth]{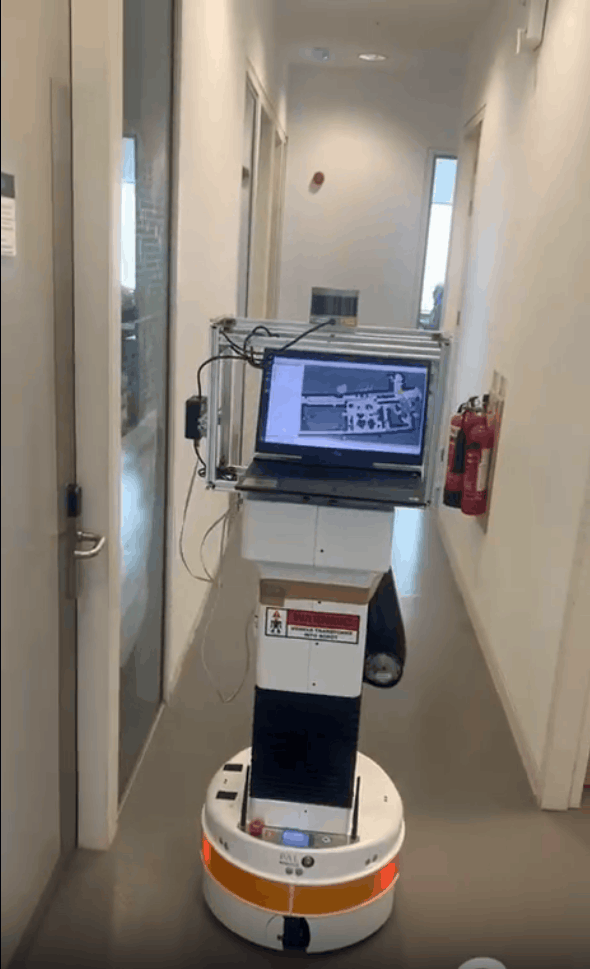}
        \end{minipage}
        
        \caption{S2---Buffet Event}
        \label{fig:exp-real-inference-buffet}
    \end{subfigure}
    \caption{Real-world experiments: (a) parameter update setup, (b) S1 corridor, and (c) S2 area. In each navigation column: [Top] Task map with Start (S) and Goal (G). [Middle] The baseline planner fails by cutting through crowded areas (blue paths). [Bottom] The causal planner anticipates congestion and takes an unobstructed detour.}
    \label{fig:exp-real-inference}
\end{figure*}
Next, we deploy the invariant causal structure on a physical TIAGo robot to assess rapid sim-to-real adaptation via light-data parameter updates and navigation performance against a non-causal baselines. Videos of the real-world experiments are publicly available online\footref{foot:website}. The real-world setting includes the S1 corridor (orange) and S2 breakout space (pink) (see the environmental map in Fig.~\ref{fig:exp-real-learning}, top). Participants were instructed to behave as naturally as possible while forming semi-static queues facing walls in S1 to replicate an academic event, and interacting while moving between mock food stations in S2.

To adapt the simulation-trained causal structure (Fig.~\ref{fig:exp-cm} left) to the real world, we updated its parameters using limited observational data: speed, location, and human trajectories captured by an onboard 3D LiDAR tracker~\cite{Yan2020a} and anonymously recorded (Fig.~\ref{fig:exp-real-learning}, middle and bottom).
%
%To adapt the simulation-trained causal structure (Fig.~\ref{fig:exp-cm} left) to the real world, we updated its parameters using limited observational data: the robot's speed and location, alongside human trajectories. These trajectories were captured by an onboard 3D LiDAR people tracker~\cite{Yan2020a} and anonymously recorded (see Fig.~\ref{fig:exp-real-learning}, middle and bottom).
%
We collected data ($2$~min per scenario at $10$~Hz) from $20$ volunteers\footnote{Study approved by the Ethics Committee of the University of Lincoln (ref.~6379), and informed consent was obtained from all participants.}. This duration aligns with prior work~\cite{castri2024experimental} showing a $120$~s window captures representative spatial dynamics while minimizing participant burden. This window provides sufficient samples for calibrating tabular CPDs over the fixed invariant structure.%This recording duration was chosen based on empirical findings from prior work on human-robot spatial interaction~\cite{castri2024experimental}, which demonstrated that an observation window of $120$~s is sufficient to capture representative spatial dynamics while minimizing participant burden. Given that our framework only estimates tabular CPDs over a fixed invariant structure, this short observational window provides adequate sample support for parameter calibration.

These observations were subsequently used to update the CPDs of our causal model,
%calibrating the synthetic simulation parameters to reflect real-world dynamics,
ensuring the robot adapted to the actual people density of the physical venue, while preserving the simulation-learnt causal graph.

\subsection{Adaptability and Sim-to-Real Transfer}
A core advantage of the proposed framework is its ability to rapidly adapt the simulation-trained model to a physical environment. A standard real-world pipeline---requiring comprehensive data collection for both causal discovery \textit{and} parameter learning---would have necessitated approximately $2$~hours of continuous data recording and $6$ minutes for causal discovery, plus $100$~ms for parameter learning. %By exploiting the invariant causal model, we avoid causal discovery and its heavy data requirement, reducing the latter to merely $2$~minutes recordings per scenario. In practice, this real-world adaptation strategy decreased the deployment time by approximately $98.4\%$.
By avoiding a new causal discovery and its heavy data requirements, data collection is reduced to just $2$~minutes per scenario, reducing deployment time by approximately $98.4\%$.

%This adaptability translates into significant time savings. A full real-world pipeline---requiring data collection sufficient for both causal discovery {\em and} parameter learning---would have necessitated approximately $2$~hours of data recording, plus $6$ minutes for causal discovery, and $100$~ms for parameter learning. By exploiting the invariant causal model, we avoid causal discovery and its heavy data requirement, reducing the latter to merely $2$~minutes recordings per scenario. In practice, this real-world adaptation strategy decreased deployment time by $\sim \mathbf{98.4\%}$.

\subsection{Navigation Performance}
%During the live experiments, participants adhered to the instructions described above, while the robot navigated. 
\subsubsection{Task Success} In both scenarios, the robot was tasked with navigating from the start node ($S$) to goal ($G$) (Figs.~\ref{fig:exp-real-inference-poster}--\ref{fig:exp-real-inference-buffet}, top). The shortest-path baseline cut directly through crowds (corridor in S1, social gathering in S2), causing navigation failures and halting near attendees before replanning (Figs.~\ref{fig:exp-real-inference-poster}--\ref{fig:exp-real-inference-buffet}, middle). %Quantitatively, treating the path waypoints as sequential subtasks, the baseline reached only a $\sim33\%$ completion rate in both tasks (3 out of 9 in S1 and 2 out of 6 in S2). 
%
%In contrast, our causal framework predicted the high congestion levels of some waypoints, ant proactively generated detours (Figs.~\ref{fig:exp-real-inference-poster}~and~\ref{fig:exp-real-inference-buffet}, bottom) reaching all planned destinations and achieving a $100\%$ completion rate (10 out of 10 waypoints in S1 and 9 out of 9 in S2), without getting stuck.
%
Quantitatively, treating the path waypoints as sequential subtasks, the baseline reached only a $\sim 33\%$ completion rate in both tasks ($3/9$ in S1 and $2/6$ in S2), planning shorter nominal paths ($21.47$~m in S1 and $19.18$~m in S2) but traveling only $15.26$~m and $6.29$~m before aborting due to congestion. In contrast, our causal framework anticipated congestion and proactively generated longer but unobstructed detours ($36.79$~m in S1 and $30.58$~m in S2 (Figs.~\ref{fig:exp-real-inference-poster}--\ref{fig:exp-real-inference-buffet}, bottom), achieving a $100\%$ completion rate ($10/10$ waypoints in S1 and $9/9$ in S2) without getting stuck. Over these short runs, physical battery depletion was negligible; evaluating long-term energy trade-offs is left for future work.%Over these single-run distances, physical battery depletion was negligible and therefore not detectable by the onboard robot status; evaluating long-term energy trade-offs across extended missions is left for future multi-trial deployments.

\subsubsection{Proxemic Compliance} The path planning improvement is also confirmed by a human-robot proxemic compliance analysis. Fig.~\ref{fig:exp-real}~[top] reports the distribution of the minimum distance between the robot and the central waypoints of the activity clusters. The baseline frequently intruded into the participants' ``Social''~($1.2$--$3.6$~m) space, particularly in S1, with occasional situations of ``Intimate'' distance. Conversely, the causal planner consistently maintained a safer and more socially acceptable ``Public'' distance~($>3.6$~m).
\begin{figure}
\vspace{-1.5mm}
    \centering
    \includegraphics[trim={0cm 0cm 0cm 0cm},clip,width=0.7\columnwidth]{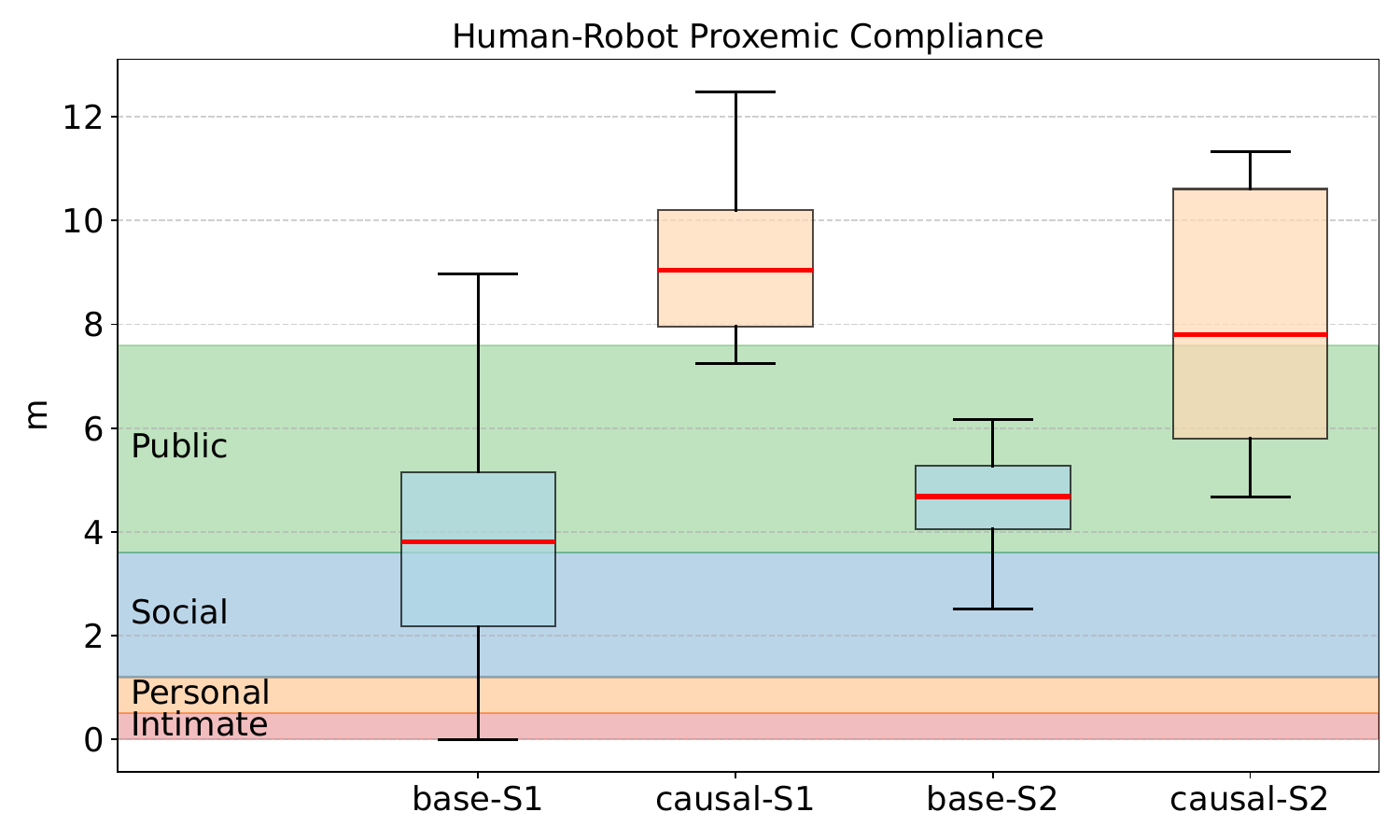}
    \includegraphics[trim={0cm 0cm 0cm 0cm}, clip, width=0.7\columnwidth]{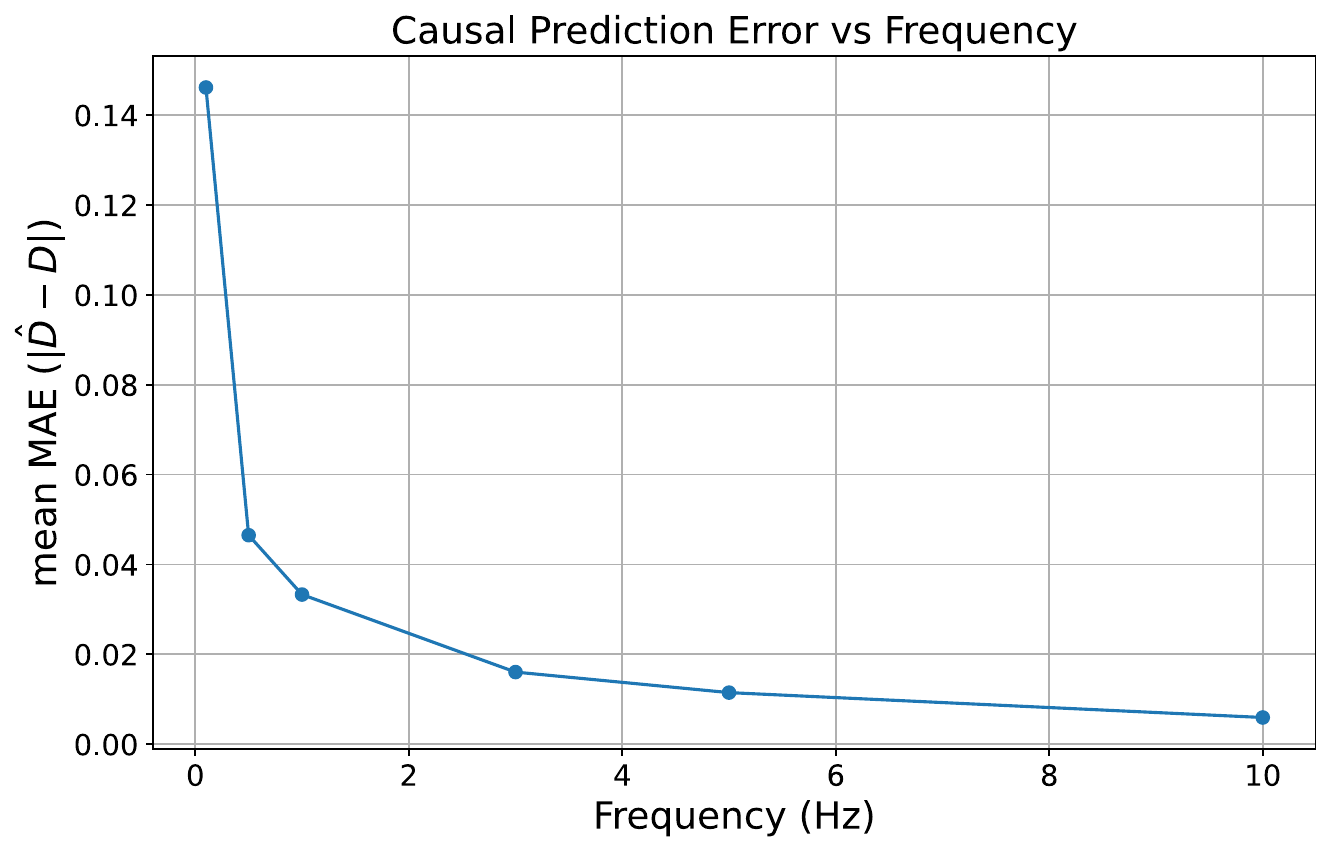}
    \caption{[Top] Real-world proxemic compliance across Hall's zones~\cite{hall1968proxemics}: Intimate, Personal, Social, and Public. [Bottom] Density prediction MAE ($\hat{D}$ vs.\ $D$) across sampling rates.}\label{fig:exp-real}
\end{figure}

\subsection{Data and Computational Efficiency}
Besides rapid adaptability, the framework demonstrates strict operational efficiency, both in terms of data requirements and computational load. To quantitatively assess the data efficiency of the parameter update, we performed a sensitivity analysis on the amount of observational data used. We subsampled the original $10$~Hz 3D LiDAR stream to collect less data at lower frequencies ($0.1, 0.5, 1, 3, 5$~Hz) and computed the MAE of the causal people density prediction~$\hat{D}$ against the ground truth~$D$. As Fig.~\ref{fig:exp-real}~[bottom] shows, the performance stabilises rapidly, reaching satisfactory levels even at low sampling rates, as typical human speeds in these scenarios are easily tracked without high-frequency sensing.

Finally, to validate the onboard viability of the framework for actual robot deployment, we measured the inference time required to query the causal engine during the physical experiments. As the system estimates battery cost and people density per topological node, inference time scales directly with map size. For our $29$-node real-world environment, the average inference time recorded across both scenarios was $\sim0.12$~s per query. This result closely aligns with the scalability analysis shown in previous simulations~\cite{castri2026causality}, confirming that the framework introduces negligible computational overhead for real-time decision-making.

\section{CONCLUSIONS}\label{sec:conclusion}
In this work, we presented the first real-world deployment of an end-to-end causal decision-making framework for autonomous mobile robots. By leveraging the invariance of high-level causal structures, a robot trained in simulation can be adapted to complex, physical human-shared environments with only a rapid, light-data parameter fine-tuning process.
Simulation and real-world experiments confirmed our research hypotheses: ({\em i})~\textit{Effectiveness}---our causal framework yields safer paths and higher task success rates than non-causal baselines by reasoning about congestion causes; 
({\em ii})~\textit{Adaptability}---leveraging structural invariance bypasses the safety hazards and data overhead of physical causal discovery, bridging the sim-to-real gap; and 
({\em iii})~\textit{Efficiency}---rapid parameter calibration cuts deployment time to two minutes while maintaining sub-second inference onboard standard robotic hardware.

While our physical trials demonstrate the feasibility and real-time execution of the framework, an extensive multi-trial evaluation across diverse real-world settings is left for future work. Future research will also extend the causal framework to handle more complex social dynamics and explore online structural adaptation under evolving causal mechanisms.

\section*{Acknowledgements}
This work has received funding from the European Union’s Horizon 2020 research and innovation programme under grant agreement No 101017274 (DARKO). GB is also supported by PNRR MUR project PE0000013-FAIR.

\bibliographystyle{unsrt}
\bibliography{biblio}

\end{document}